\documentclass[letterpaper]{article} 
\usepackage{aaai24}  
\usepackage{times}  
\usepackage{helvet}  
\usepackage{courier}  
\usepackage[hyphens]{url}  
\usepackage{graphicx} 
\usepackage{natbib}  
\usepackage{caption} 
\usepackage{algorithm}
\usepackage{algorithmic}
\usepackage{multirow}
\usepackage{array}
\usepackage{dsfont}

\usepackage{amssymb}
\usepackage{amsmath}
\usepackage[table]{xcolor}
\usepackage{newfloat}
\usepackage{listings}
\DeclareCaptionStyle{ruled}{labelfont=normalfont,labelsep=colon,strut=off} 
\floatstyle{ruled}
\newfloat{listing}{tb}{lst}{}
\floatname{listing}{Listing}
\title{When mitigating bias is unfair: multiplicity and arbitrariness \\ in algorithmic group fairness}
\author{
    Natasa Krco\textsuperscript{\rm 1}\equalcontrib\thanks{Work conducted during an internship at AXA},
    Thibault Laugel\textsuperscript{\rm 2, 3}\equalcontrib,\\
    Vincent Grari\textsuperscript{\rm 2,3,4},
    Jean-Michel Loubes\textsuperscript{\rm 5},
    Marcin Detyniecki\textsuperscript{\rm 2,3,6} 
}
\affiliations{
    \textsuperscript{\rm 1}Imperial College, London, United Kingdom \quad
    \textsuperscript{\rm 2}AXA, Paris, France \quad
    \textsuperscript{\rm 3}TRAIL, LIP6, Sorbonne Université, Paris, France \quad
    \textsuperscript{\rm 4}Stanford University, United States \quad
    \textsuperscript{\rm 5}Institut de Mathématiques de Toulouse, Université Paul Sabatier, Toulouse, France \quad
    \textsuperscript{\rm 6}Polish Academy of Science, IBS PAN, Warsaw, Poland \\
    n.krco23@imperial.ac.uk, thibault.laugel@axa.com
}

\begin{document}

\nocopyright

\setcounter{secnumdepth}{0} 

%



\maketitle

\begin{abstract}


Most research on fair machine learning has prioritized optimizing criteria such as Demographic Parity and Equalized Odds. Despite these efforts, there remains a limited understanding of how different bias mitigation strategies affect individual predictions and whether they introduce arbitrariness into the debiasing process. This paper addresses these gaps by exploring whether models that achieve comparable fairness and accuracy metrics impact the same individuals and mitigate bias in a consistent manner. We introduce the FRAME (FaiRness Arbitrariness and Multiplicity Evaluation) framework, which evaluates bias mitigation through five dimensions: Impact Size (how many people were affected), Change Direction (positive versus negative changes), Decision Rates (impact on models’ acceptance rates), Affected Subpopulations (who was affected), and Neglected Subpopulations (where unfairness persists). This framework is intended to help practitioners understand the impacts of debiasing processes and make better-informed decisions regarding model selection. Applying FRAME to various bias mitigation approaches across key datasets allows us to exhibit significant differences in the behaviors of debiasing methods. These findings highlight the limitations of current fairness criteria and the inherent arbitrariness in the debiasing process.

\end{abstract}




\section{Introduction}

As Machine Learning becomes an increasingly integral and frequent part of systems used in high-stakes applications that directly impact people, the concern about the potential risks and harms these systems may carry grows as well. Regulations such as the GDPR or the European AI Act are established to ensure that AI systems do not cause harm or infringe the rights of those affected by them. One of the risks that must be accounted for is that of automated systems creating or enforcing inequality and discrimination. Consequently, the field of algorithmic fairness that aims at studying the unwanted biases these systems may create or amplify, has grown exponentially over the recent years~\cite{jobin2019global}.
Yet, as a complex ethical and philosophical concept, fairness has been proven difficult to formalize as a machine learning objective. Numerous, sometimes conflicting, definitions and ways to quantify bias have been proposed (see e.g. the relevant surveys:~\citet{Romei2013AMS,del2020review, mehrabi2021survey, hort2022survey, besse2022survey}). 

Among them, group fairness aims to evaluate bias by calculating and comparing global scores between sensitive groups, e.g. acceptance rates (\emph{Demographic Parity}) or True Positive and Negative rates (\emph{Equalized Odds}).
These criteria have already been the target of criticism, questioning their purely technical perspective. 
By turning a social problem into a mathematical optimization one, works in algorithmic fairness have thus been shown to be failing to account for several key aspects of the bias problem, generally suffering from being too simplistic~\cite{binns2020apparent, vzliobaite2011handling} and ignoring any social context~\cite{selbst2019fairness,wachter2021why}.

\begin{figure}[t]
    \centering
    \includegraphics[width=1.\linewidth]{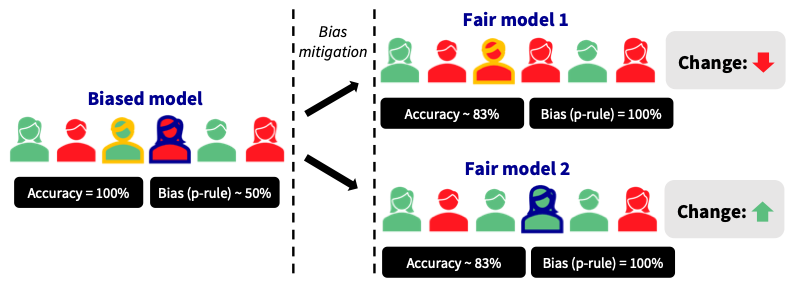}
    \caption{Illustration of multiplicity in the debiasing process: replacing an existing (biased) model with a fair one is an underspecified problem. Several models (here, two are shown) achieve the same accuracy and fairness scores by adopting drastically different strategies.}
    \label{fig:multiplicity-debiasing-illustration}
\end{figure}

In parallel, recent works have highlighted the by-nature underspecified aspect of machine learning tasks and the resulting problems that may arise~\cite{d2020underspecification,marx2020predictive,renard2021understanding}. 
Multiple models can be learned on the same data and achieve similar predictive performance despite being very different, raising the question of the arbitrariness of the model selection process as a whole, and its social and legal implications ~\cite{black2022model,creel2022algorithmic}. This raises the question of to what extent multiple models might also achieve the same group fairness and accuracy scores but differ in behaviour, and what would be the consequences of this multiplicity in the \emph{debiasing process}, i.e. the process of replacing a biased model with a fair one. To illustrate this idea, we describe a simple scenario in Figure~\ref{fig:multiplicity-debiasing-illustration}. Considering a model used for a given task (here called "the biased model"), bias mitigation using two different mitigation methods leads to two candidate fair models. The first one (top row) achieves fairness (as measured by $p\%-rule$, the higher the fairer
), by reducing demographic disparity. This is accomplished by altering the status of an accepted (green) male applicant to a refused (red) one. 
The second one achieves the same levels of accuracy and fairness by turning a rejected woman into an accepted one. 
This scenario illustrates what we identify as \emph{multiplicity in the debiasing process}. As they reach the same performances on both criteria, the two fair models are often considered equivalent, despite achieving their objectives in opposite ways. The selection of one model over another is often done in a blind fashion~\cite{balayn2023fairness}, heavily relying on global metrics such as accuracy and $p\%-rule$. 
We argue that this multiplicity therefore introduces significant arbitrariness into the decision process that should be accounted for in order to ensure the fairness of the debiasing process as a whole, as well as the final model itself.

Unfortunately, this arbitrariness is made harder to control by the opaque aspect of group bias mitigation. Indeed, although a few works focus on achieving fairness using transparent models~\cite{aivodji2019learning}, bias mitigation methods often rely on opaque models such as deep learning~\cite{zhang2018adversarial,adel2019one}, ensemble~\cite{grari2019fair} models, or complex projections of the data to reduce the influence of the sensitive attribute~\cite{zemel2013learning}. But besides the resulting fair, Pareto efficient, model often being a black box, more generally the {bias mitigation process} itself is not transparent. As a result, the practice of enforcing algorithmic fairness for a given real-world application is generally a completely intractable process and its impacts, both at the global and individual levels, remain unobserved and misunderstood.

In this paper, we therefore aim to formulate a new argument against the blind optimization of group fairness criteria through the lens of model multiplicity.
After empirically showing its prevalence, we discuss its potentially harmful consequences, arguing that independently of the context, settling for any group fairness metrics will generally lead to some notion of arbitrariness, hence unfairness, in the process.
To circumvent these issues, we argue that it is crucial to make debiasing a transparent process. For this purpose, we propose the FRAME (FaiRness Arbitrariness and Multiplcity Evaluation) framework 
to help characterize bias mitigation. FRAME relies on 5 dimensions, framed as questions and quantitative tests, whose answers are designed to help a ML practitioner understand the impacts of a considered debiasing method and make a more informed decision when selecting a model over the others.
After introducing FRAME, we apply it to several bias mitigation approaches across some of the most-represented datasets of the fairness literature. 
Besides drawing insights on the differences between bias mitigation approaches, this allows us to illustrate how this framework can be leveraged by practitioners to understand the differences and impacts of debiasing, and thus help mitigate the associated arbitrariness. Finally, we discuss research directions that are raised by our empirical findings.



\section{Problem Statement}
\label{sec:background}

In this section, we first examine the different ways in which the concept of fairness is understood and approached in current literature, and give an overview of related work. Finally, the motivations behind our work are presented.

\subsection{Group Fairness}

In this work, we consider binary classification problems with binary sensitive attributes: a dataset $D$ with $n$ samples $(x_i, s_i, y_i)_{i=1}^{n}$ is used to train a model $f$. An instance $i$ is represented by $x_i\in \mathbf{R}^m$ - its feature vector, $s_i\in \{0,1\}$ - its binary sensitive attribute, and $y_i\in \{0,1\}$ - its binary target. The population is split by the sensitive variable into a \textit{privileged} ($s_i=1$) and a \textit{disadvantaged} subgroup ($s_i=0$), and the classification $\hat{y}_i$ is into a positive ($\hat{y}_i=1$) and a negative ($\hat{y}_i=0$) class (desirable and undesirable outcome). 
We focus on \textit{fairness through awareness}~\cite{dwork2012fairness} - approaches in which the sensitive attribute is accessible during the training, test and debiasing process, and in particular \textit{group} fairness, which  requires that a chosen criterion such as false positive or false negative rate is equal across subgroups of a population.
However, group fairness is not at all straight-forward to define - defining or choosing a criterion to characterize it is a research topic in itself, and there exists therefore a highly diverse group of proposed interpretations~\cite{hort2022survey}. We discuss here the two most commonly used criteria, that will be the focus of this paper.


\paragraph{Demographic Parity}

A classifier satisfies Demographic Parity if the prediction $\hat{y}$ is independent from sensitive attribute $s$. In the case of a two class classification, this is equivalent to 
    \begin{equation}
        P[\hat{Y}=1 | S=0] = P[\hat{Y}=1 | S=1]
    \end{equation}
The intuition behind this criterion is that subgroups by sensitive attribute will have an equal chance of a positive outcome. For measuring fairness in terms of Demographic Parity, the \textbf{p\%-rule} metric is often used:
    \begin{equation}
        p\%-rule = min( \frac{P[\hat{Y} = 1 | S=1]}{P[\hat{Y} = 1 | S=0]}, \frac{P[\hat{Y} = 1 | S=0]}{P[\hat{Y} = 1 | S=1]} )
    \end{equation}
A 100\% \textbf{p\%-rule} value signifies perfect fairness, indicating no predictive disparity between the groups.


\paragraph{Equalized Odds} A classifier satisfies the Equalized Odds criterion if the prediction \(\hat{Y}\) is conditionally independent of the sensitive attribute \(S\) given the actual outcome \(Y\)~\cite{hardt2016equality}:
\begin{align}
       P[\hat{Y}=1 | S=0, Y=\textit{y}] = P[\hat{Y}=1 | S=1, Y=\textit{y}], \\ \forall y \in \{ 0,1\} \nonumber
\end{align}
This criterion extends beyond Demographic Parity by incorporating the ground truth into fairness assessments.


In doing so, it motivates making changes to the predictions that align with the target labels. We use \textbf{Disparate Mistreatement (DM)}, introduced in~\cite{zafar2017fairness} to measure  
equalized odds as the addition to the two following terms:
    \begin{align}
        D_{TPR} = |TPR_{S=1} - TPR_{S=0}| \\
        D_{FPR} = |FPR_{S=1} - FPR_{S=0}| 
    \end{align}
    where
    \begin{align}
        TPR_{S=s}  = P[\hat{Y}=1 | Y=1, S=s], \text{    } s \in \{ 0,1 \}   \\
    FPR_{S=s} = P[\hat{Y}=1 | Y=0, S=s], \text{    }  s \in \{ 0,1 \} 
    \end{align}
    are the true positive and false positive rates per group.
The closer the values of $D_{TPR}$ and $D_{TNR}$ to 0, the lower the degree of disparate mistreatment of the classifier. In the rest of the paper, we refer to DM as the sum of $D_{TPR}$ and $D_{TNR}$. 

\subsection{Group Fairness Algorithms}\label{Group Fairness Algorithms}
Numerous strategies have been developed to optimize the fairness metrics discussed in the preceding section. Although various categorization schemes exist, this work concentrates on the framework introduced by~\citet{Romei2013AMS}, which categorizes the fairness enforcement strategies into three different families:
\begin{itemize}
    \item \textbf{Pre-processing}: This first category of methods focus on debiasing the training data itself.
    Relying on methods such as strategic sampling \cite{manerba2022investigating} and latent space projection~\cite{zemel2013learning,kamiran2009classifying,gordaliza2019obtaining}, a fair dataset is either generated, or sampled from the existing one, with the hope that classifiers trained on this dataset would be less biased.   
    \item \textbf{In-processing}: In this case, the goal is to modify the training process in such a way that the new resulting model will be fair. Multiple approaches to this have been proposed, such as adding a fairness term to the learning objective~\cite{kamishma2012prejudice, agarwal2018reductions}, or using adversarial learning~\cite{zhang2018adversarial} to remove information about the sensitive attribute either from the final predictions, or from an intermediate layer of the classifier to be trained~\cite{xu2018fairgan, adel2019one, grari2021learning}.
    \item \textbf{Post-processing}: Contrary to previous categories where a new fair machine learning model is learned from scratch,
    post-processing methods modify the predictions of an existing, biased, model with the aim of making enough changes so that a fairness threshold is satisfied. 
    This can be achieved, for instance, by assigning different thresholds to each subgroup (such as in~\cite{hardt2016equality}), or by flipping predictions the model is not confident on, as seen in~\cite{kamiran2012roc}.
\end{itemize}

\subsection{The Disagreement Problem in Group Fairness}

\label{sec:background-motivations}


\begin{figure}[t]
    \centering
    \includegraphics[width=0.95\linewidth]{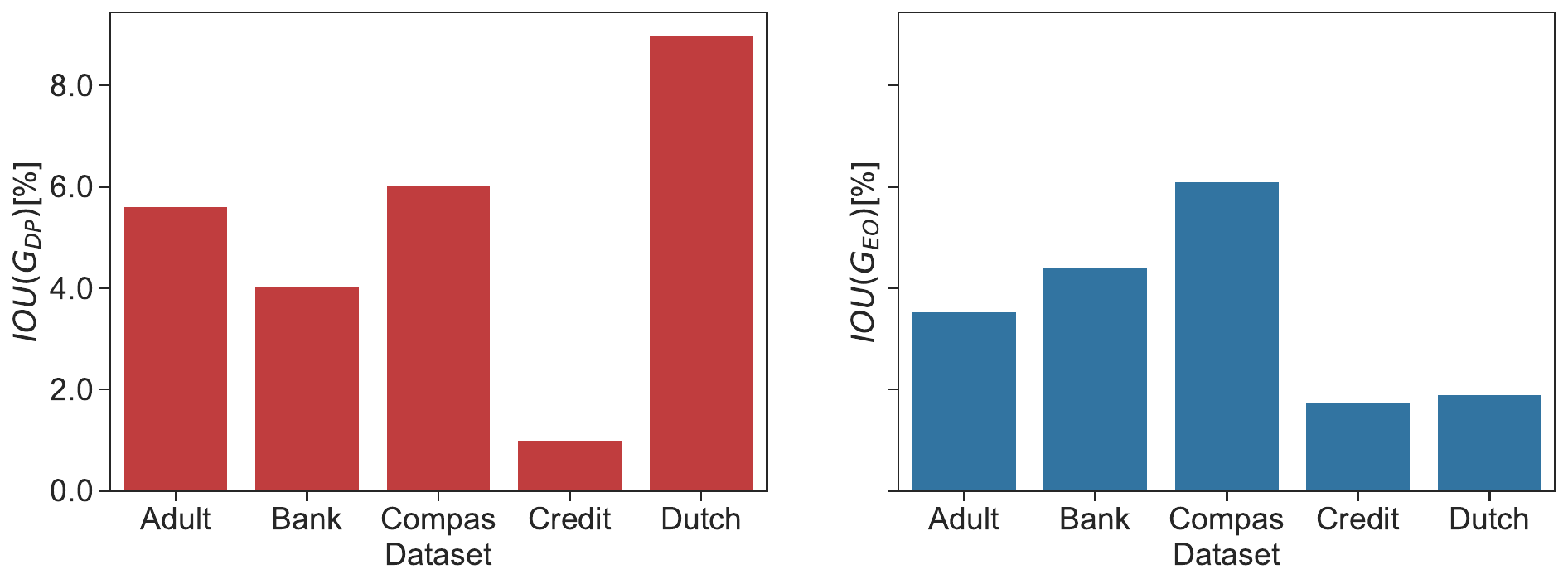}
    \caption{How similar are the predictions of fair models with similar performance? IOU values for the sets of instances targeted by the considered debiasing algorithms, for Demographic Parity (left) and Equalized Odds (right).}
    \label{fig:q2_iou}
\end{figure}

The "disagreement problem", also referred to as "predictive multiplicity", "Roshomon effect", or "model discrepancy"~\cite{marx2020predictive,renard2021understanding,black2022model} is the phenomenon observed when several machine learning models reach the same predictive performance, albeit being very different.
Generally identified as a consequence of the underspecification of machine learning problems, a natural solution consists in trying to specify more learning tasks~\cite{damour2022underspecification}, in which case fairness constraints can be seen as an answer to this issue.

Therefore, the question naturally arises of whether simply imposing new fairness constraints as an objective is enough to better guarantee the functioning of the final model as a whole, and avoid unwanted behaviors. Concurrently to our work, \citet{long2024individual} theoretically show that multiplicity is actually worsened when enforcing fairness constraints. 
Below, we propose an experiment to further  answer this question empirically and illustrate these disagreements. 

\subsubsection{Empirical Evidence}

For this purpose, we use 5 tabular datasets taken from the algorithmic fairness literature and described later in the paper. For each of them, we consider a model $f: \mathcal{X}\rightarrow \mathcal{Y}=\{0,1\}$ trained on dataset $X_{train} \subset \mathcal{X}$ to solely optimize some predictive performance metric (here accuracy). For the sake of clarity, we call~$f$ in the context of this work the \emph{biased} model.
We also consider a set of models $\mathcal{G}=\{g_i\}_i$, trained on the same dataset $X_{train}$, with an additional fairness objective, given a protected binary attribute $S$. The models $g_i$ are trained to optimize simultaneously accuracy and a fairness criterion (here either for Demographic Parity or Equalized Odds), using algorithms that are described later in the paper. All models are trained to achieve similar accuracy and fairness results (exact results in Table~\ref{tab:metrics} in Appendix), with these criteria being evaluated over a validation set $X_{val}\subset \mathcal{X}$.
To analyse the disagreement between the models $g_i$, we focus on the subpopulations \emph{treated} 
in the debiasing process at test time, i.e. the individuals from $X_{val}$ whose predictions differ between the biased model $f$ and a fairer model $g_i$. We introduce the following notation for these instances:
$$ \Delta_{g_i} = \{ x \in \mathcal{X} | f(x) \neq g_i(x) \}  $$
We then  assess the amplitude of the disagreement between the models of $\mathcal{G}$ by measuring the IOU between the subpopulations treated by the models: $
IOU(\mathcal{G}) = \frac{ | \bigcap_{g\in \mathcal{G}} \Delta_g| }{ | \bigcup_{g\in \mathcal{G}} \Delta_g  |} 
$.

The results are shown in Figure \ref{fig:q2_iou}. 
Across all datasets, the overlap between the individuals targeted by each method is consistently below $9\%$, highlighting how different the bias mitigation process is for the various approaches. 
This number goes even down to lower than $2\%$ in the case of the DP methods for the Credit dataset.

Going further, we additionally examine how stable the set of targeted instances for one method~$\Delta_g$ is across multiple runs. We therefore assess the 
percentage of instances changed across 10 training runs for each specific debiased model $g \in \mathcal{G}$. 
Table~\ref{tab:stability-dp} illustrates the percentage of instances changed in every run for various fairness methods across different datasets for DP methods (results for EO can be found in Table~\ref{tab:stability-eo} in Appendix), as well as the average number of instances changed for reference. Again, we observe differences in the behaviors of the methods: because it is deterministic, ROC is fully stable; on the other hand, LFR and the in-processing methods   
demonstrate less stability, as the portion of individuals targeted in every run is low.
Besides classical stability concerns, this further motivates questioning the fairness of debiasing while relying only on global metrics - even within the same method, an individual may or may not be affected depending on the run. 

\begin{table}[t]
   \centering
   \resizebox{\columnwidth}{!}
   {\begin{tabular}{|c|c|c|c|c|c|c|}
       \hline
       Dataset & Metric & Adversarial (DP) & ROC & LFR & TO (DP) & EGR (DP) \\ \hline \hline
       \multirow{2}{*}{Adult}& Mean $\%$ of instances changed &11.55& 11.17 &11.66&8.57& 18.89\\
       & $\%$ instances changed in every run &4.15&11.17&2.4&8.49&0.06 \\ \hline \hline
       \multirow{2}{*}{Bank}& Mean $\%$ of instances changed &8.20&1.65 &8.77&3.79&9.37 \\
       & $\%$ instances changed in every run &1.18&1.65&7.28&2.72& 0.16\\ \hline \hline
       \multirow{2}{*}{COMPAS}& Mean $\%$ of instances changed &16.54& 13.64&24.51&15.98& 31.12 \\
       & $\%$ instances changed in every run &1.51&13.64&3.25&13.52&29.70 \\ \hline \hline
       \multirow{2}{*}{Credit}& Mean $\%$ of instances changed &7.43& 0.42&6.74&1.03& 18.41\\
       & $\%$ instances changed in every run &3.1&0.42&4.21&0.64&0.03 \\ \hline \hline
       \multirow{2}{*}{Dutch}& Mean $\%$ of instances changed &20.71& 47.66 &25.75&18.20&27.35 \\
       & $\%$ instances changed in every run &12.96& 47.66 &4.57&17.32& 0.01\\ \hline
   \end{tabular}}
   \caption{Stability of the instances targeted by DP methods across 10 runs. All results are in $\%$ of $\mathcal{X}_{val}$.}
   \label{tab:stability-dp}
\end{table}


Overall, these results highlight the prevalance of multiplicity in debiasing. This further underlines the influence of the choice of debiasing strategy and randomness, as despite achieving similar scores, the affected populations vary significantly.


\subsubsection{Unfairness from Arbitrariness}
Just like in a traditional supervised learning setting, multiplicity in debiasing may not necessarily be seen as a problem per se. However, we argue that blindly selecting one debiased model over the others is harmful to the true objective of fairness. 

First, selecting a model arbitrarily among several seemingly equivalent candidates results in an arbitrary selection of the treated population as well: in Fig.~\ref{fig:multiplicity-debiasing-illustration}, Fair Model 1 results in one man having his outcome changed from desirable (in the biased setting) to undesirable. In the absence of any justification, this choice (this individual rather than other men) may be perceived as unfair due to its seeming arbitrariness~\cite{black2022model}. This is even more significant as mitigation strategies have been shown to unevenly affect individual probability rankings~\cite{goethals2024beyond}. Second, as previously discussed by~\citet{lipton2018does}, debiasing methods may cause the classifier to predict based on other, hidden criteria, in order to satisfy fairness constraints, resulting in unobserved or illogical discrimination. More recently, \citet{grari2023fairness} discuss how optimizing group fairness metrics may not affect all subpopulations equally - e.g. older people may be ignored when mitigating a gender bias. In cases such as this, only certain parts of the population actually benefit from the debiasing, leaving other parts to continue to suffer from unequal treatment.

Finally, despite being seemingly equivalent, these models impact the overall population in very different manners. Looking back at the example of Fig~\ref{fig:multiplicity-debiasing-illustration}, although the two fair models have identical accuracy and fairness scores, they have different acceptance rates: $1/3$ (Fair Model 1), and $2/3$ (Fair Model 2). Therefore, selecting one model over the other results not only in changes for the treated subpopulations, but also for the population overall. For Fair Model 1, this may be perceived as another evidence of the "levelling down" effect~\cite{mittelstadt2023unfairness}, stating that enforcing fairness may negatively impact the whole population. 

In such scenarios, we therefore argue that despite the models meeting their mathematical expectations, enforcing fairness fails to achieve its true purpose: rather than ensuring fairer decisions, it merely shifts the threat of unfairness to another problem, stemming from the arbitrariness of the decisions and the underspecification of the fairness problem.
This generally aligns with existing critiques on the optimization culture of the field of algorithmic fairness~\cite{selbst2019fairness,balayn2023fairness,wachter2021fairness}, pointing out its possible adverse effects and therefore advocating for an examination of the fairness problem from a more socio-technical perspective. 
Yet, existing investigations on the behavior of fairness models generally focus on quantitative assessments of their performance under various situations, such as their sensitivity to the training data~\cite{friedler2019comparative}, robustness under perturbations and attacks~\cite{nanda2021fairness,kamp2021robustness}, or preservation of the biased model's logics~\cite{manerba2022investigating,goethals2024beyond}. 
While also valuable, the main limitation of such benchmarks is that they only adopt a global quantitative perspective on these models, which can hardly be leveraged in a legal or social reflection. Moreover, they generally overlook the aforementioned complexities raised by predictive multiplicity and arbitrariness, effectively masking these critical issues.
In contrast, the goal of this work is to provide tools that delve into the impacts of the debiasing process both at the global and individual levels. Hence, by doing so, we intend to enable practitioners to gain a better understanding of its implications overall, and thus make better-informed decisions when enforcing fairness. 


\section{FRAME: a framework for FaiRness Arbitrariness and Multiplicity Evaluation 
}
\label{sec:methodology}

\begin{table*}[]
    \centering
    \begin{tabular}{|l|c|c|c|}
         \hline
          Dimension & Description  & Inputs required & Impact scale\\
          \hline
          \hline
         [D1] Impact size & How many people were affected? & $X_{val}, f, g_i$ & Global  \\
         \hline
         [D2] Change direction  & How many positive changes vs. negative changes? & $X_{val}, S, f, g_i$ & Individual \\
         \hline
         [D3] Decision rates & What was the impact on the models' acceptance rates? &  $X_{val}, g_i$ & Global \\
         \hline
         [D4] Affected subpopulations & Who was affected by the debiasing process? & $X_{val}, S, f, g_i$ & Individual  \\
         \hline
         [D5] Neglected subpopulations & Is the model still unfair for some subpopulations? & $X_{val},S, g_i$ & Global\\
         \hline 
    \end{tabular}
    \caption{Description of the FRAME Framework.}
    \label{tab:framework-description}
\end{table*}

To address the concerns described in the previous section, we propose to address the problem of multiplicity through the lens of transparency. As fair models may achieve similar performances using very different strategies, we propose tests to characterize the impacts of these strategies both at the individual, group and global levels. 
The overall framework, FRAME, is described below.

\subsection{General Presentation}

FRAME relies on 5 dimensions, framed as questions and gathered in Table~\ref{tab:framework-description}.
To use FRAME, the same context as the one described in Section~\ref{sec:background} is considered: it therefore requires access to a "biased" model $f$ (optimizing a predictive performance criterion), a replacement candidate $g_i$ (jointly optimizing a predictive performance and a unique fairness criterion), and a validation dataset $X_{val}$. Both models only need to be accessible as oracles - the framework is, therefore, completely model-agnostic, both for the biased model and the fair one.
To answer each question, a test is proposed. It is important to note that FRAME is not intended as a benchmark - there is no clear "correct" answer to any of the proposed questions. The main goal of FRAME is to make the debiasing process more transparent, and help practitioners better understand the differences between the biased and fair models.

\subsection{Assessment Dimensions}
\label{sec:background-researchquestions}

Below, we describe each dimension, what it captures, and what risks it can help mitigate.

\paragraph{DIMENSION 1: Impact size} 
Rather than "debiasing a model", most works in algorithmic fairness focus on proposing to replace the biased model with a completely new one. This raises the question of the remaining similarity between the original model and its replacement, that we set to answer in this first dimension by answering the question: "How many people were affected by the debiasing process?"
In contexts where fairness is to be enforced when a known model is already in production, having models that do not differ much from the previous one could indeed be desirable: although biased, the former model may have been validated by experts, or undergone additional tests supporting its behavior. 
Additionally, this question is also connected to issues of consistency: shifting decisions may negatively impact customers' trust~\cite{burgeno2020impact}. 
Yet, very few works proposing fair models focus on this issue.
To answer the question, we measure the proportion $\frac{|\Delta_{g}|}{|X_{val}|}$ of people targeted by each debiasing method, i.e. people for whom the fair model predicts a different class than the biased model.

\paragraph{DIMENSION 2: Change direction} 
Rather than measuring the size of the population affected by the debiasing process, the objective of this second dimension is to characterize better \emph{how} these individuals are affected, as all changes are not equivalent for the individuals. In particular, we focus, for the affected populations, on the \emph{direction} of the predicted outcome change: we thus consider $g(x)>f(x)$ (resp. $g(x)<f(x)$) to be a \textit{positive} (resp.  \textit{negative}) change. The importance of this dimension can be visualized in the illustrative example of Fig.~\ref{fig:multiplicity-debiasing-illustration}, with the Fair Model 1 negatively affecting a man and the Fair Model 2 positively affecting a woman. For example, in the context of a model for hiring decisions, someone having their job application rejected by the biased model, but accepted by the fair one is considered a \textit{positive} difference. In this context, it can be expected that negative changes may hurt user trust even more than positive ones. 

\paragraph{DIMENSION 3: Decision rates}
Beyond impacts at the individual level, the behavior captured by the previous dimension is also connected to global consequences on the model. A debiasing approach relying on positive changes will evidently not have the same effect on the model's acceptance rate as one relying on negative changes. To put the observations made in the previous dimension into perspective, we look at the final acceptance rates $\mathbb{E}(g_i(X_{val}))$ (for DP methods) and FPR and FNR (for EO methods) obtained by the fair models. In the example of Figure~\ref{fig:multiplicity-debiasing-illustration}, we see that despite having identical performances, the two fair models achieve completely difference acceptance rates ($1/3$ and $2/3$). As a result, the final decision model could approach the task for which it is used in vastly different ways, depending on the chosen debiasing method. 

\paragraph{DIMENSION 4: Affected subpopulations}
Assessing the impacts of the debiasing process at the individual level also requires understanding which populations are targeted by each debiasing method. The goal of this dimension is therefore to gain insight into the typically opaque strategies adopted by each considered approach to achieve fairness, and potentially identify hidden, undesirable behaviors. 
For this purpose, we propose to leverage XAI techniques~\citep{carvalho2019machine}, in particular post-hoc methods, aiming to generate explanations for the predictions of a trained classifier. However, following our methodology, rather than explaining directly classifiers $f$ or $g_i$, we aim here to describe the set $\Delta_g$. Concretely, this can be done by applying XAI tools to a fictional binary decision model: $x\rightarrow \mathds{1}_{x\in \Delta_g}$. As post-hoc explainability methods rely on various assumptions, they are therefore prone to disagreements~\cite{krishna2022disagreement}. Hence, the choice of the explainability method should be done carefully, while keeping in mind that there is not a single correct way to conduct this analysis. Promising strategies could thus include training and visualizing a decision tree predicting the labels $\mathds{1}_{x\in \Delta_g}$, clustering analysis of the sets $\Delta_g$, or using local XAI methods such as SHAP~\cite{lundberg2017unified}. In this paper, we propose instead to leverage Partial Least Squares Discriminant Analysis~\cite{wold2001pls}, which aims to train a linear classification model in a lower dimension, interpretable, space. It thus provides the upside of finding the most relevant dimensions to disambiguate the populations targeted by each method, which can then be easily visualized.

\paragraph{DIMENSION 5: Neglected subpopulations}
Previous works~\cite{hashimoto2018fairness} have argued that traditional fairness notions could be combined with a "Rawlsian" notion of fairness, following which a system is as unfair as its worst-off group. This has been formalized as a generalization problem~\cite{hashimoto2018fairness,ferry2023improving}, as it has been shown that not all subpopulations are guaranteed to benefit from group debiasing~\cite{grari2023fairness}. Small, underrepresented subpopulations may indeed be ignored during the training process, and as a result, a model with good global fairness performance may still be locally unfair. 
This dimension therefore aims to assess whether fairness is enforced in subgroups of interest, defined in terms of features of $X$.
Specifically, we propose to calculate fairness metrics (p\%-rule and Disparate Mistreatment) across different splits of the dataset, in order to determine whether the fair models are equally fair across these subgroups.
These subgroups are context-dependant and should be defined by the practitioners based on the specific dataset and task they work with. Besides ensuring that no subpopulation of interest is neglected when enforcing bias, measuring this notion of local fairness aligns with the idea that fairness should be particularly enforced for \emph{comparable} subgroups~\cite{wachter2021why,vzliobaite2011handling}.


\section{Applying FRAME}

In the rest of the paper, we apply FRAME to several datasets and bias mitigation approaches, commonly used in the algorithmic fairness literature.
Besides showcasing the value of our framework, the objective of this analysis is to derive insights about common fairness approaches and shed light on potential adverse effects of bias mitigation.
In this section, we describe the experimental setup considered for this analysis.
\label{sec:experimental-protocol}

\subsection{Datasets}
\label{sec:datasets}
We apply our framework on five tabular datasets commonly used in the algorithmic fairness literature~\cite{lequy202survey}. We consider a binary sensitive attribute and binary class label for each dataset, and use the same $50\%$ train-test split of a dataset across all (fair and unfair) models: Adult~\cite{misc_adult_2}, Dutch~\cite{dutchdataset}, COMPAS~\cite{angwin2016machine}, Bank~\cite{bankdataset} and Credit~\cite{creditdataset}.
Description of the target and sensitive variables for these datasets can be found in Table~\ref{tab:datasets} in Appendix.

\subsection{Debiasing Strategies}\label{sec:fairness-strategies}

To demonstrate the use of FRAME, we choose approaches representative from each of the three categories described in the "Problem Statement" section: preprocessing (LFR~\cite{zemel2013learning}), inprocessing (EGR~\cite{agarwal2018reductions} and Adversarial Debiasing~\cite{zhang2018adversarial}), and postprocessing (ROC~\cite{kamiran2009classifying} and TO~\cite{hardt2016equality}). We focus on methods optimizing Demographic Parity on one side (LFR, EGR, Adversarial Debiasing, ROC and TO), and Equalized Odds on the other side (Adversarial Debiasing, EGR and TO).
When available, we use existing implementations of methods from the IBM AIF360 \cite{aif360} and Microsoft Fairlearn~\cite{fairlearn} libraries.
The specific methods along with their categories and metrics they optimize are listed in Table \ref{tab:methods} in Appendix. Models obtained by applying these methods are later referenced in the paper as the \emph{fair models}.

\subsection{Exeperimental Protocol}

\begin{figure}[h]
    \centering
    \includegraphics[width = 0.5\textwidth]{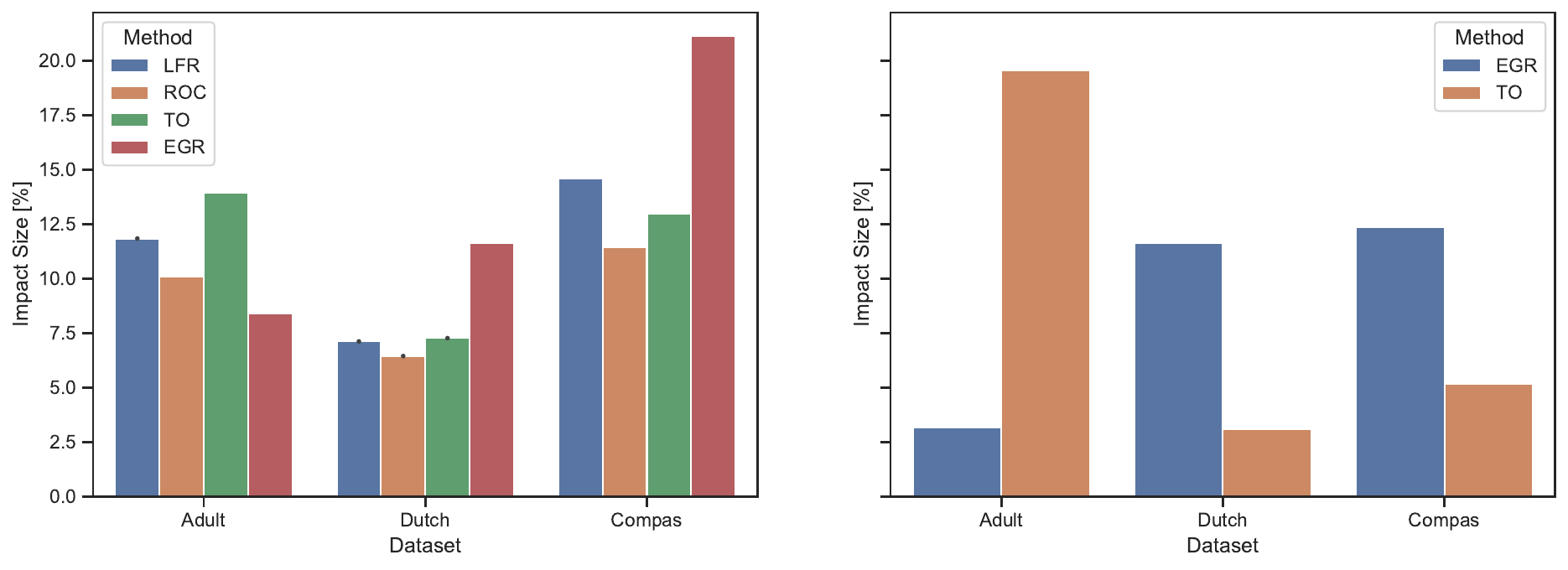}
    \caption{Impact size: Percentage of instances impacted. Left: Demographic Parity methods. Right: Equalized Odds methods. Results are in $\%$ of the test set.}
    \label{fig:xgb_d1}
\end{figure}

\begin{figure}[h]
    \centering
    \includegraphics[width=0.23\textwidth]{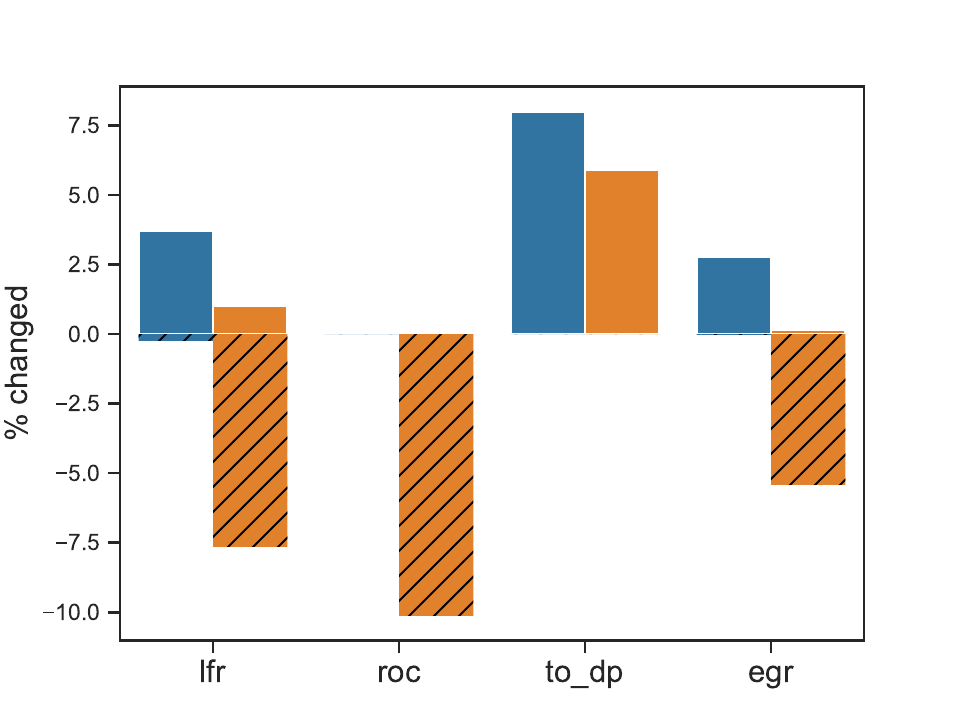}
    \includegraphics[width=0.23\textwidth]{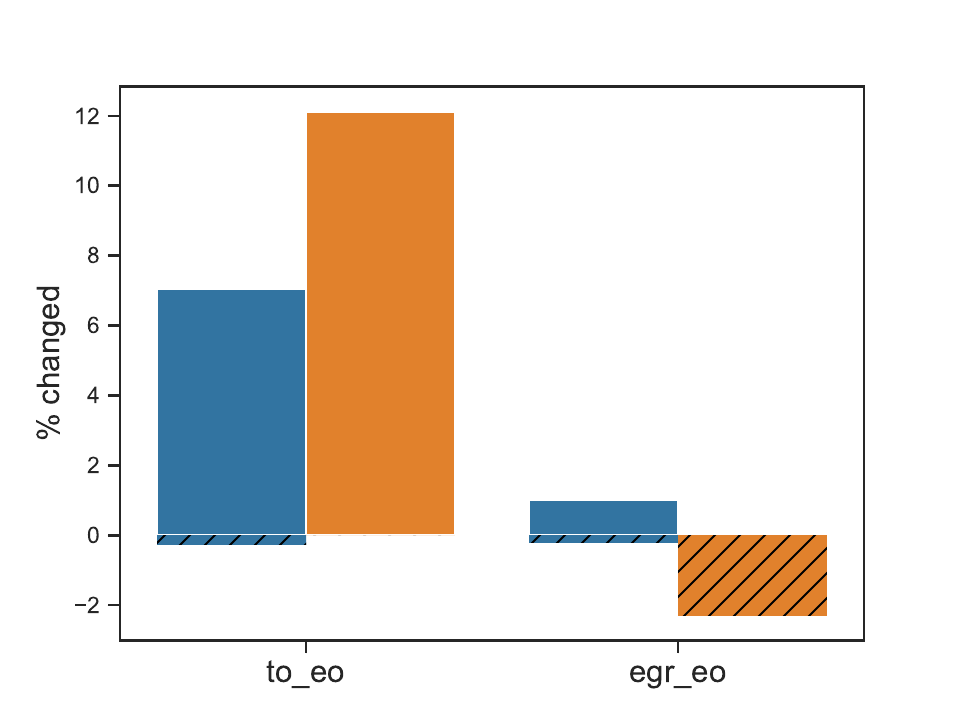}
    \caption{Change direction for GBX trained on Adult dataset}
    \label{fig:xgb_d2_adult}
\end{figure}

\begin{figure}
    \centering
    \includegraphics[width=0.22\textwidth]{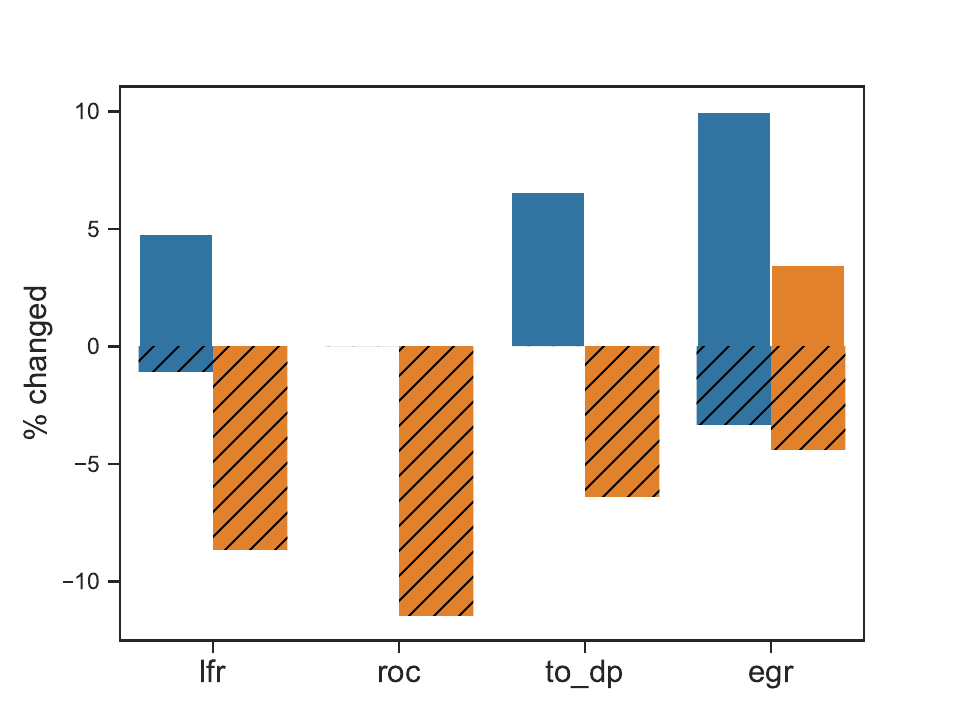}
    \includegraphics[width=0.22\textwidth]{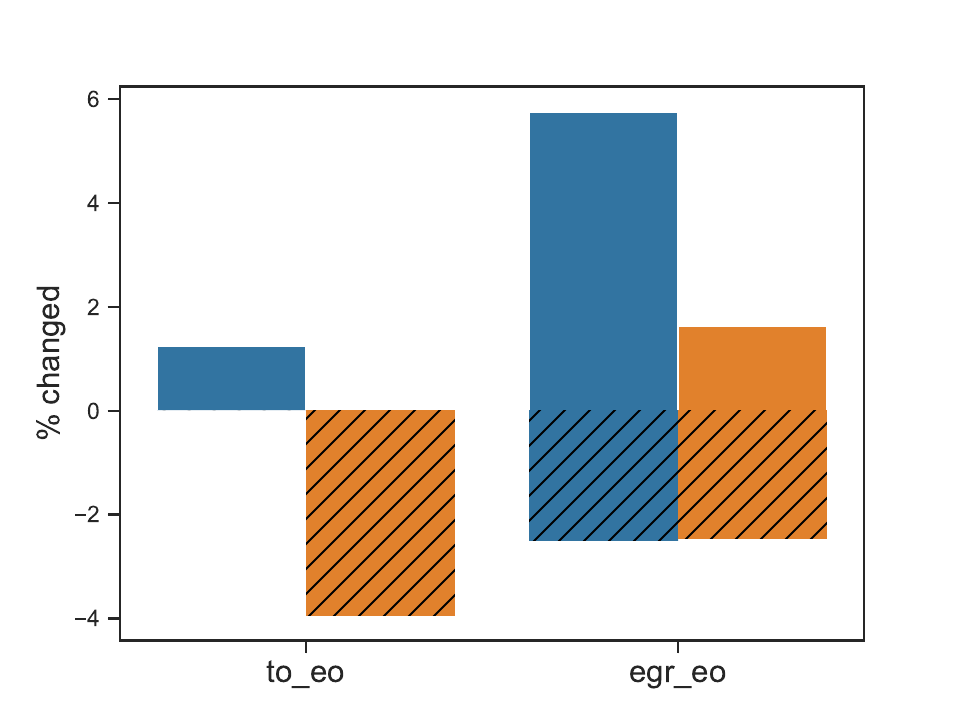}
    \caption{Change direction for GBX trained on Compas dataset}
    \label{fig:xgb_d2_compas}
\end{figure}

\begin{figure}
    \centering
    \includegraphics[width=0.22\textwidth]{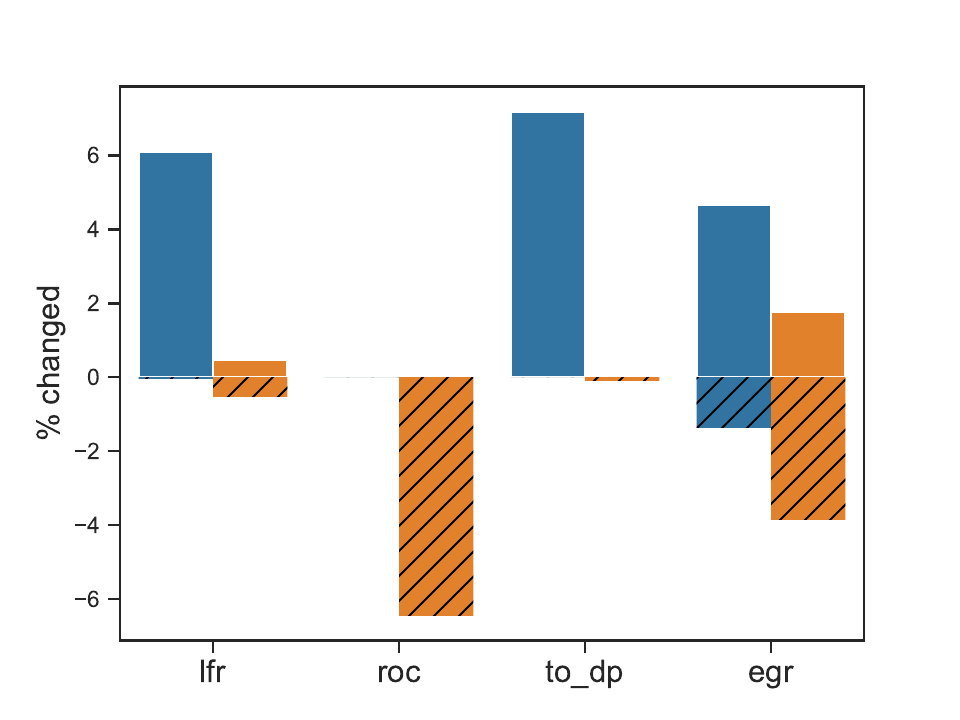}
    \includegraphics[width=0.22\textwidth]{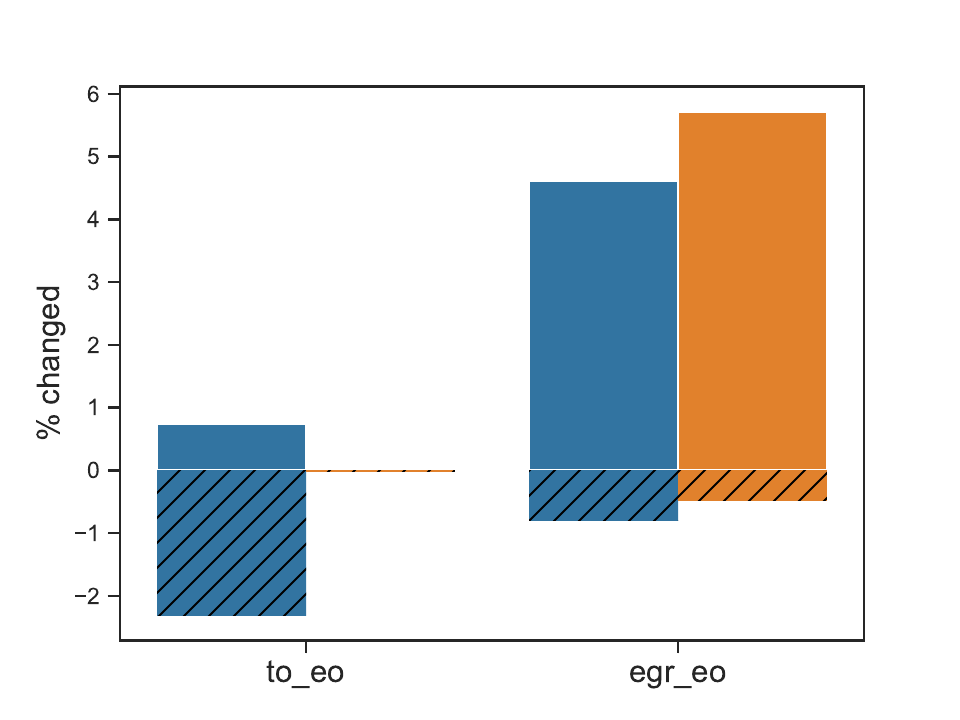}
    \caption{Change direction for GBX trained on Dutch dataset}
    \label{fig:xgb_d2_dutch}
\end{figure}

\paragraph{Training a biased model.} 
Our proposed methodology relies on comparing the predictions of fair models to the ones of a prior biased model.
For our biased model, we use a fully connected neural network optimizing only a performance criterion (binary cross-entropy). 
We choose a neural network both due to its performance on the chosen datasets, and its compatibility with some of the chosen bias mitigation strategies, in particular adverarial debiasing methods. 
However, this choice does not impact the results, and other types of classifiers (e.g. tree-based methods, logistic regression...) could have been considered for this biased model. 
For the sake of completeness, we have reported in Appendix performance metrics (Table~\ref{tab:metrics_xgb}) and the results for experiments concerning D1 (Figure~\ref{fig:xgb_d1}) and D2 (Figures~\ref{fig:xgb_d2_adult}, \ref{fig:xgb_d2_dutch}, \ref{fig:xgb_d2_compas}) using an XGBoost predictor as the biased model for three datasets.
For each dataset train-test split iteration, we train a single biased model, that we use as a reference for comparing fair models.

\paragraph{Training fair models with similar performances.}
Once the biased model is trained, we train the bias mitigation approaches described earlier. 
\begin{itemize}
    \item \textbf{Pre-processing (LFR~\cite{zemel2013learning})} We apply the LFR method to the data, and train a model with an identical architecture to that of the biased model on the pre-processed data to obtain a fair model.
    \item \textbf{In-processing: (EGR~\cite{agarwal2018reductions} and Adversarial Debiasing~\cite{zhang2018adversarial})}.
    We train these in-processing methods to optimize accuracy and fairness metrics.
    \item \textbf{Post-processing (ROC~\cite{kamiran2009classifying} and TO~\cite{hardt2016equality})} As both post-processing methods are applied to a model after it is trained, we apply them directly to the biased model's predictions.
\end{itemize}

All fair models are trained on the same training set as the biased model. As one of the goals of this study is to raise awareness on the unobserved behaviors of bias mitigation strategies, where possible, we aim for fair models that achieve similar scores, both in terms of predictive performance and fairness level. We therefore choose the hyperparameters of the presented algorithms to achieve comparable predictive and fairness scores (e.g. $\lambda$ hyperparameter balancing the accuracy and fairness losses in Adversarial Debiasing~\cite{zhang2018adversarial}). The performance metrics obtained on the considered test datasets are shown in Table~\ref{tab:metrics} in Appendix, while the values chosen for the hyperparameters can be found in Appendix (cf. Table~\ref{tab:hyperparams}).
Once trained, we apply FRAME to understand the behaviors of these models on the validation set.


\section{Results}


\subsection{D1: Impact size}
\label{sec:question1}

\begin{figure}[t]
    \centering
    \includegraphics[width=\linewidth]{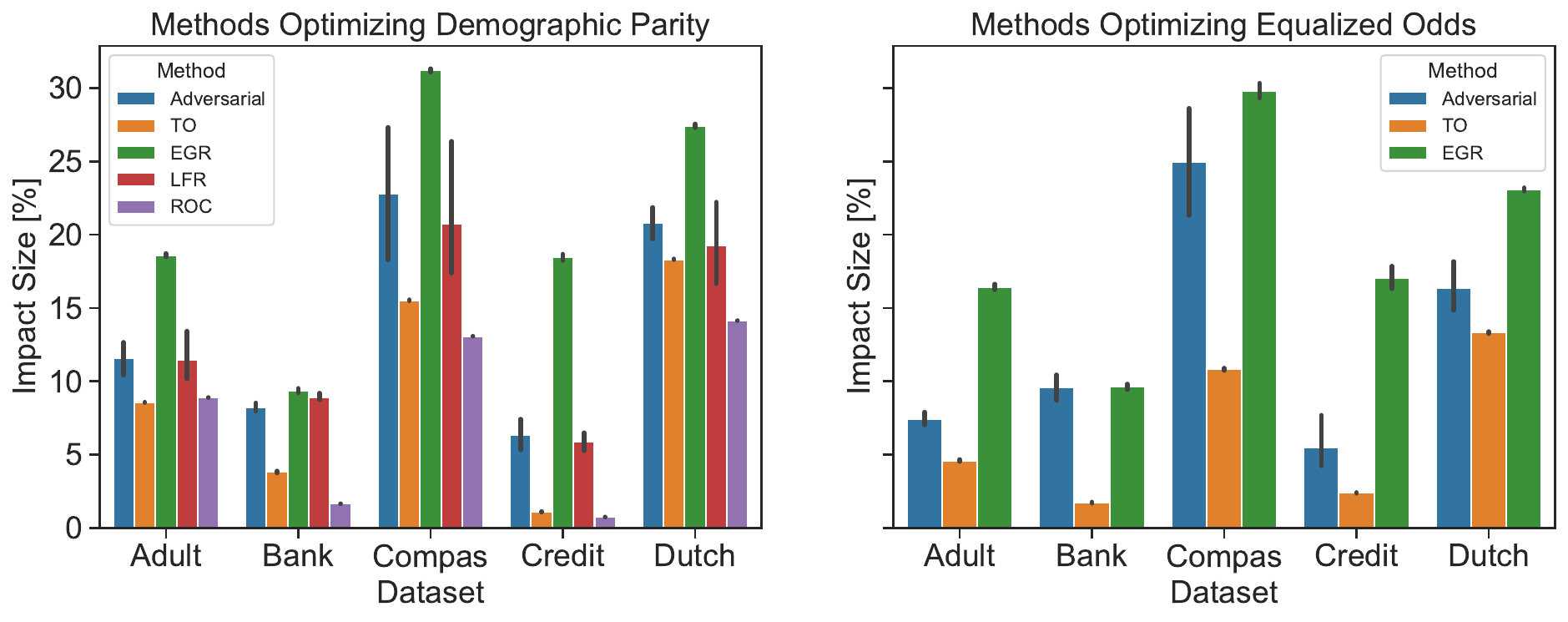}
    \caption{Impact size (D1): Average and standard deviation number of instances impacted, for 10 runs of each method, for each dataset. Left: Demographic Parity methods. Right: Equalized Odds methods. Results are in $\%$ of the test set.}
    \label{fig:d1}
\end{figure}

We first take a look at the impact size of each bias mitigation method, measuring the proportion of the test set for which predictions by the biased and the fair models disagree: $\frac{|\Delta_{g}|}{|X_{val}|}$. Results are shown in Figure \ref{fig:d1} both for methods optimizing Demographic Parity (left), and those optimizing Equalized Odds (right). We observe clear differences between methods in impact size within each dataset, with Demographic Parity methods differing by up to $15\%$ (Compas).  
Across all datasets, post-processing approaches (TO and ROC) seem much more efficient than their pre-processing and in-processing counterparts. This does not seem surprising as they directly build on the biased classifier rather than proposing a completely new model, and are therefore less inclined to adopt a drastically different strategy. Additionally, we note by looking at the error bars that these results are not equally stable for all methods. In particular, the deep network-based methods Adversarial and LFR appear to be the most unstable.



Keeping in mind that the all the presented fair models perform equally well in terms of accuracy and fairness metrics, on a given dataset, some methods thus seem more "efficient" (in terms of number of differences) in achieving fairness. On the other hand, others are possibly making sub-optimal changes to the predictions, with therefore a higher risk of affecting the behavior of the model in unanticipated ways. This may be an issue in contexts where updating a deployed (biased) model with a fair one that differs more than it is necessary (up to $31\%$ for EGR on Compas (DP)) means losing the possibly verified, controlled knowledge it carried.
Similarly, it represents a source of inconsistency of decisions over time, which can be problematic, for instance in the context of customers facing different algorithmic decisions depending on which model is in production. In contexts where these questions represent crucial challenges, more "efficient" methods such as the post-processing ones should thus be favored.




\subsection{D2: Change Direction}

\begin{figure}
    \centering
    \includegraphics[width=0.47\linewidth]{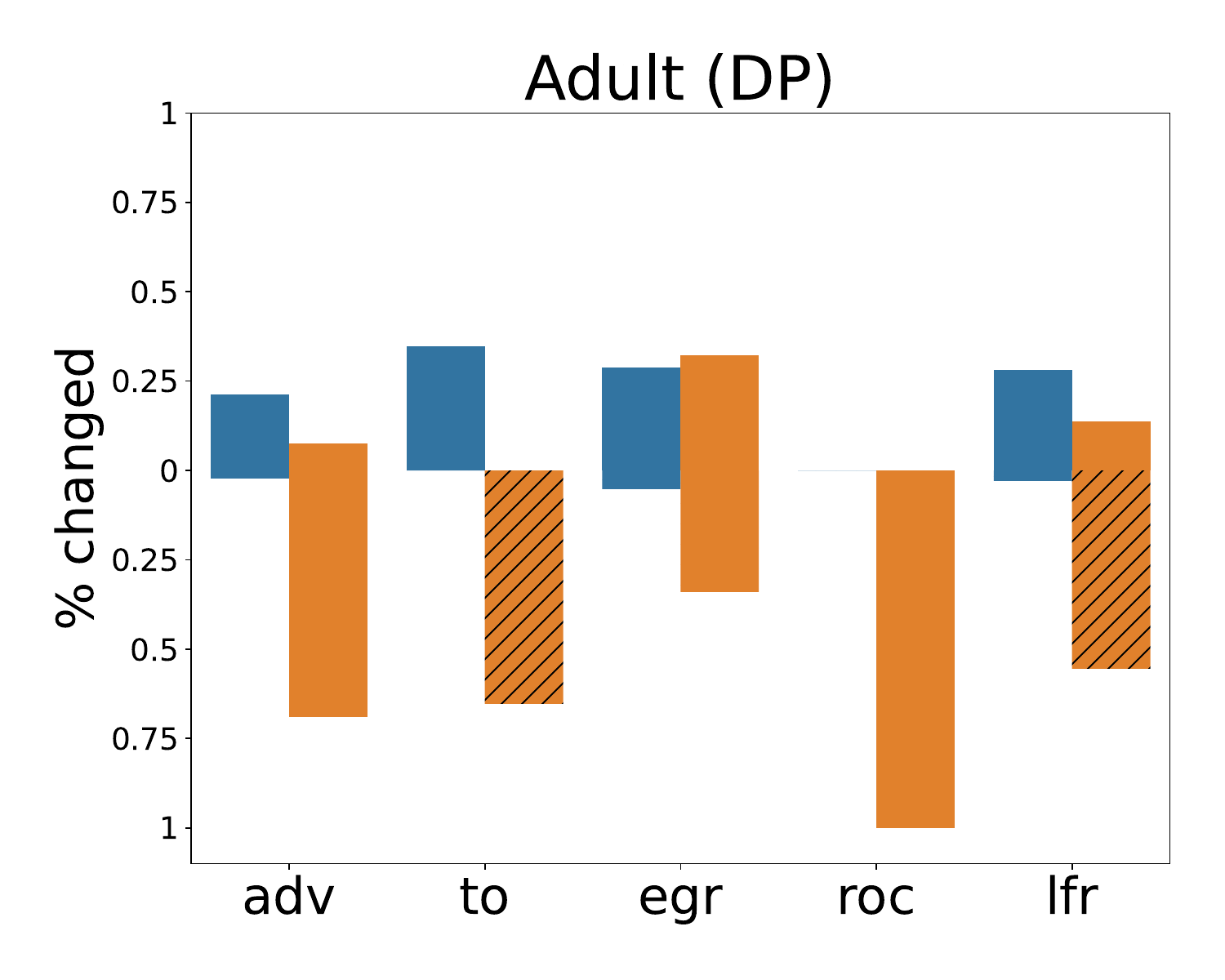}
    \includegraphics[width=0.47\linewidth]{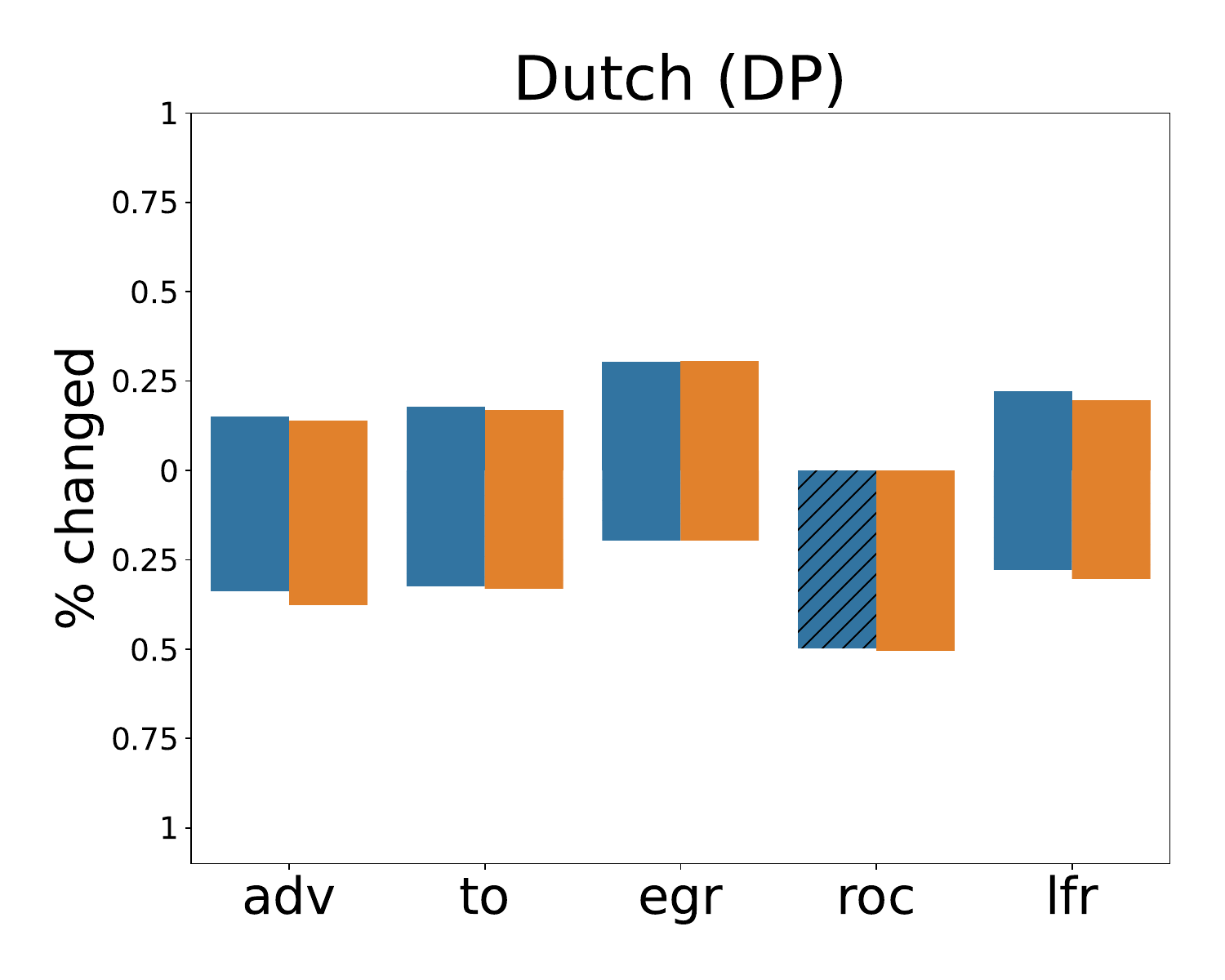}
    \includegraphics[width=0.47\linewidth]{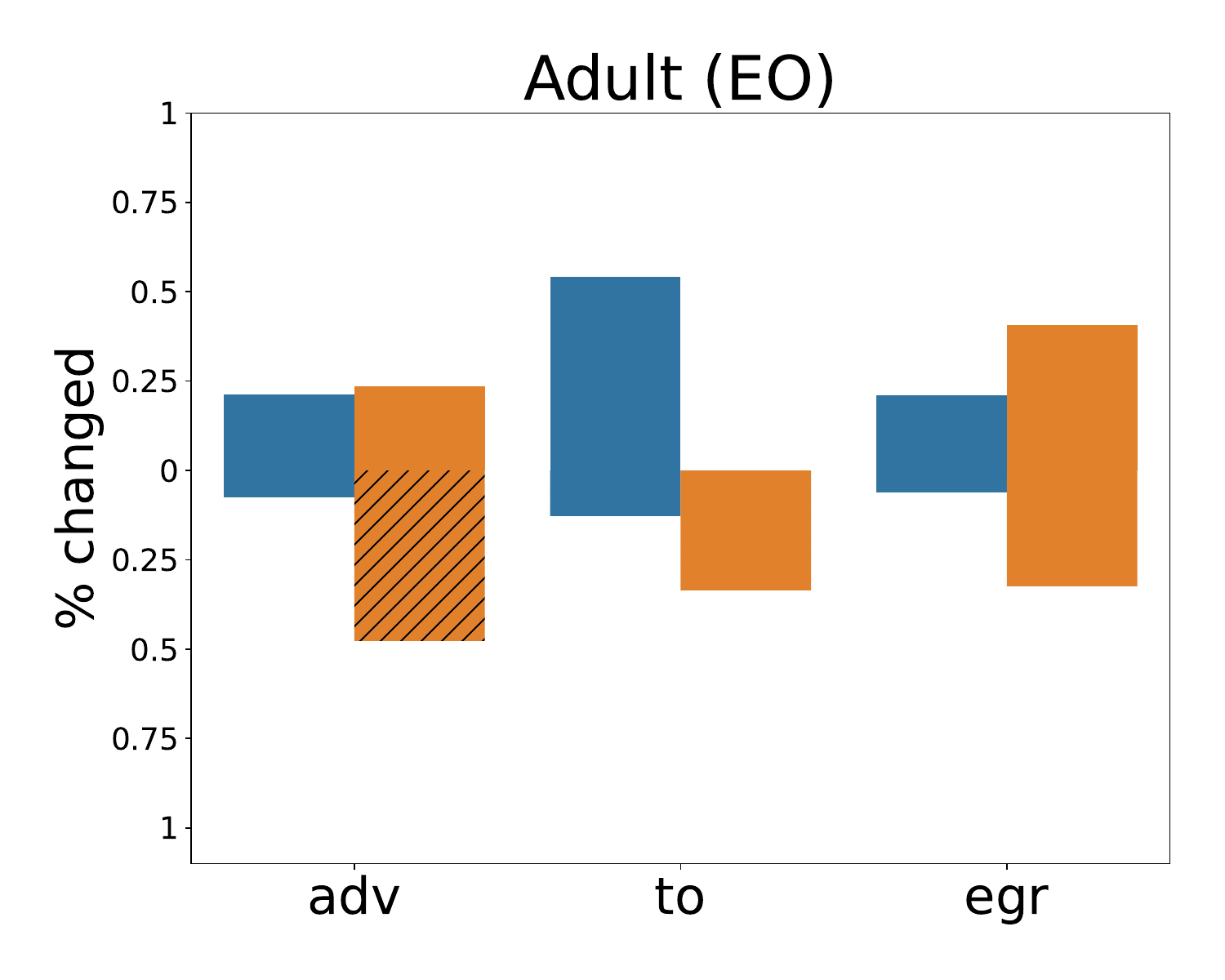}
    \includegraphics[width=0.47\linewidth]{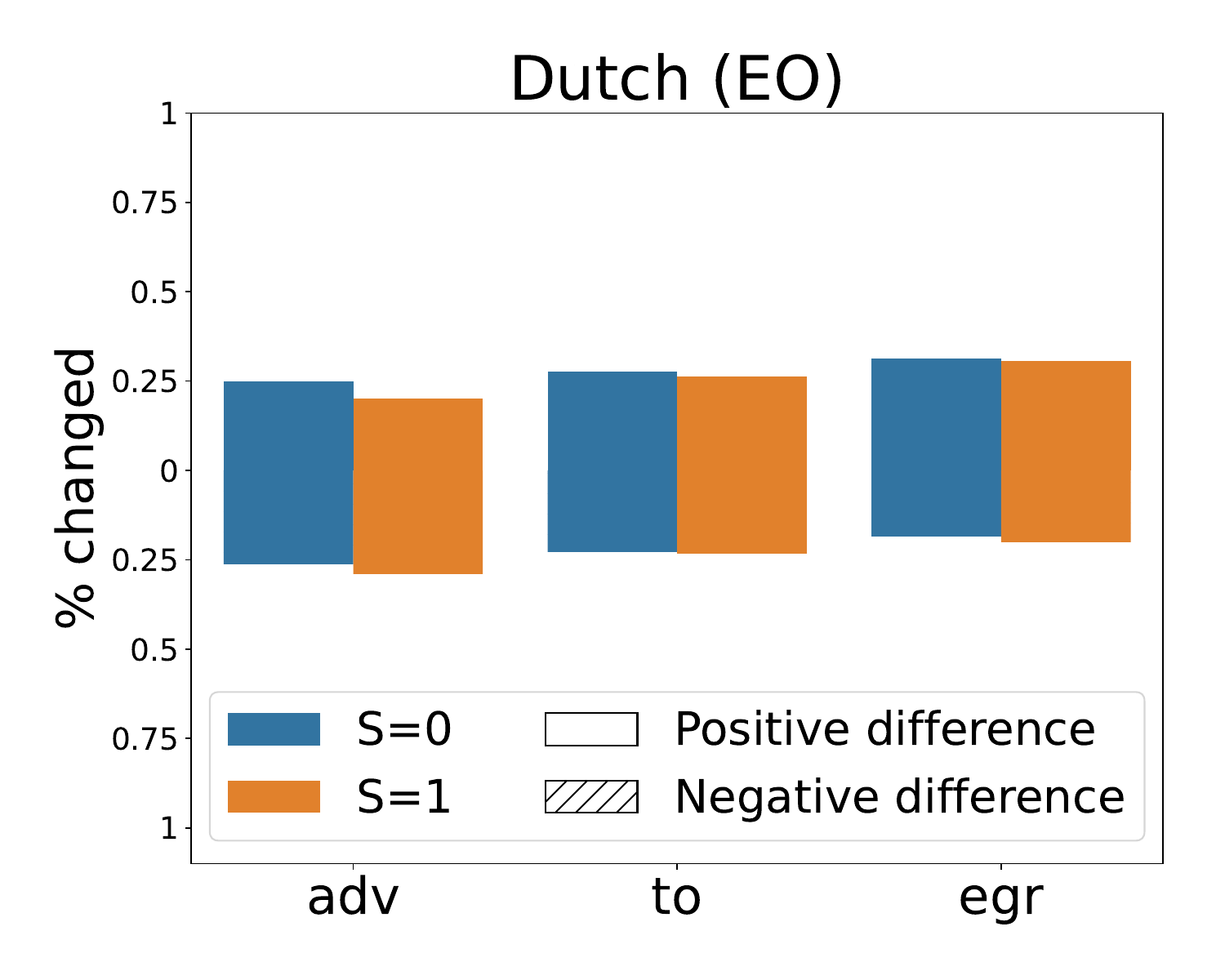}
    \caption{Change direction (D2): distribution of the observed differences along the sensitive groups and the change direction, for the Adult (left) and Dutch (right) datasets. The bar heights correspond to the proportion, out of the individuals affected by a change, of each sensitive group that got affected by a positive (full rectangles) or negative (hatched rectangles) difference.}
    \label{fig:direction-change}
\end{figure}

This second dimension aims at shedding more light on \emph{how} these fair models differ.
We therefore propose in this section to further investigate the behaviors of these methods in terms of sensitive groups and direction changes, to characterize the strategies they adopt to achieve higher fairness scores.



The distribution of the instances of~$\Delta_g$ along these two dimensions for Adult and Dutch can be seen in Figure~\ref{fig:direction-change}, for Demographic Parity and Equalized Odds.
Two observations can be derived from these results - first, the methods have significantly different strategies for enforcing fairness, with some heavily focusing on placing more privileged people in the undesirable class (e.g. all methods except EGR for the Adult dataset (DP)), while others increase the number of disadvantaged individuals in the desirable class (e.g. TO for the Adult dataset (TO)). Others have more nuanced strategies, leveraging both positive and negative differences to achieve fairness (e.g. most methods for the Dutch dataset).

An interesting second observation is that the strategies some of these methods adopt may cause counter-intuitive differences. Given the considered setting and for a given decision rate, we call an \emph{intuitive} difference to increase fairness a positive difference to an individual from the minority group ($S=0$), and a negative difference to an individual from the positive group. For the Adult dataset shown here, we notice that for some methods, particularly EGR and LFR for Demographic Parity for instance, 
over respectively $37\%$ and $15\%$ of the induced differences seemingly do not contribute to increasing Demographic Parity in a mathematical sense at all.
Yet, this obviously does not mean that these changes are useless per se: it is likely that some of these differences are consequences of larger changes in the model that are necessary to maintain a high accuracy score. 
Nevertheless, this further strengthens the idea developed in the previous experiment that some methods may be more efficient, in terms of number of changes, to reach comparable performance and fairness scores: again, the cost-efficiency of post-processing methods can easily be visualized. %

\subsection{D3: Decision Rates}

\begin{figure}[t]
    \centering
    \includegraphics[width=0.48\linewidth]{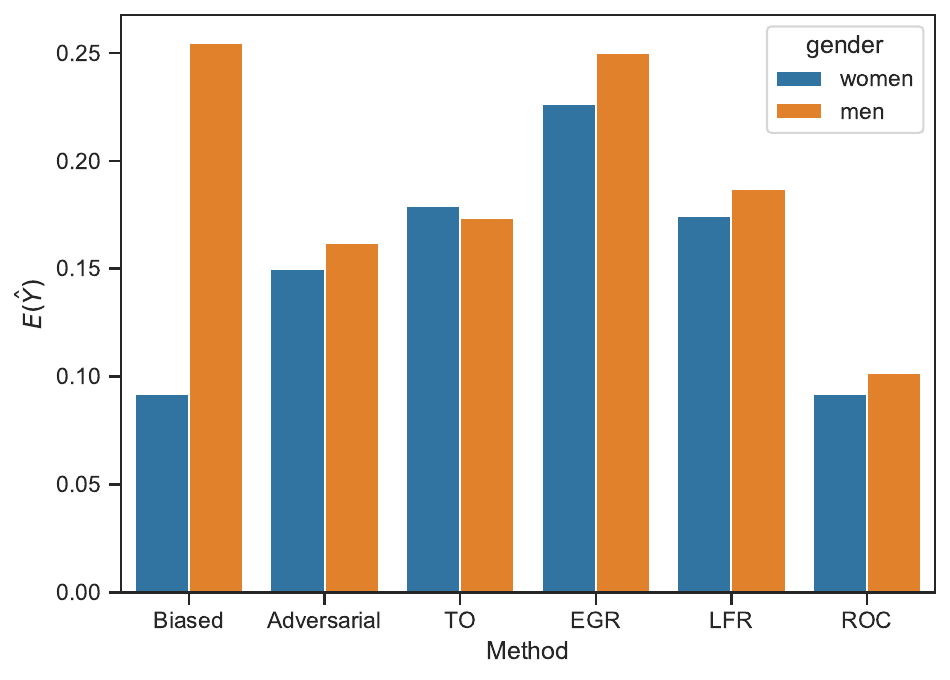}
    \includegraphics[width=0.48\linewidth]{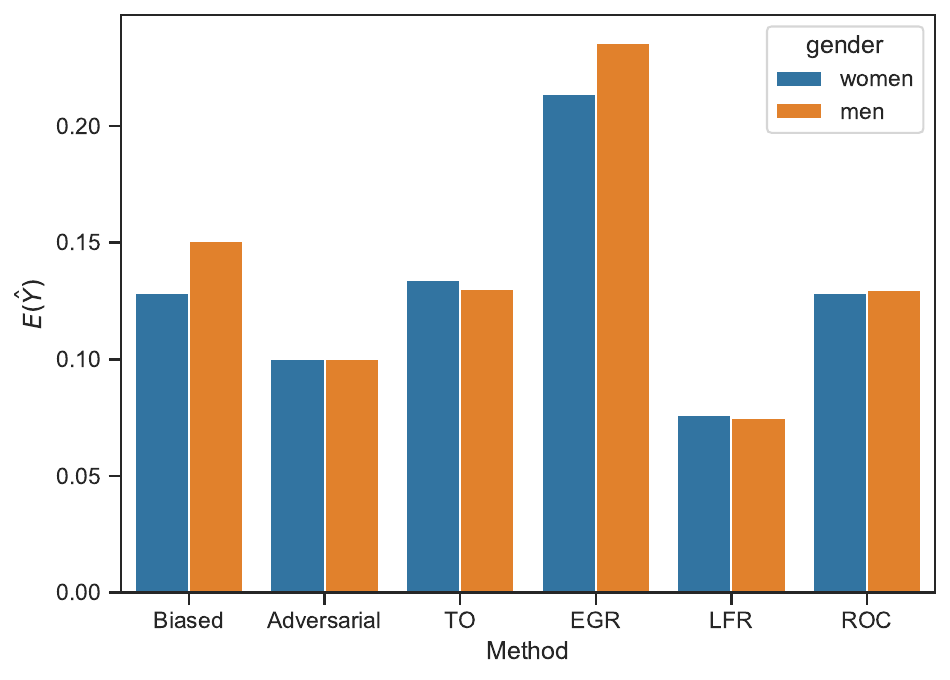}
    \caption{Final decision rates on Demographic Parity (D3): $\mathbb{E}(\hat{Y})$ values achieved by bias mitigation models for each sensitive group  of the Adult (left) and Credit (right) datasets).}
    \label{fig:q3-EY}
\end{figure}

\begin{figure}[t]
    \centering
    \includegraphics[width=\linewidth]{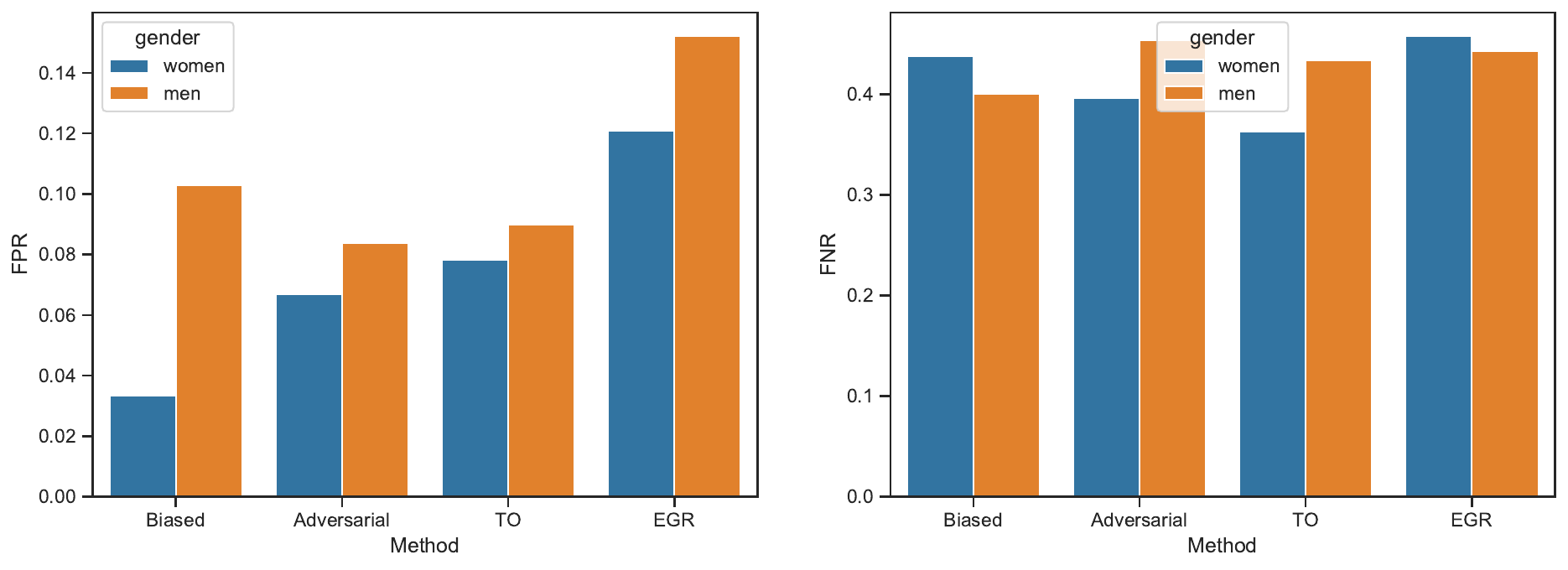}
    \includegraphics[width=\linewidth]{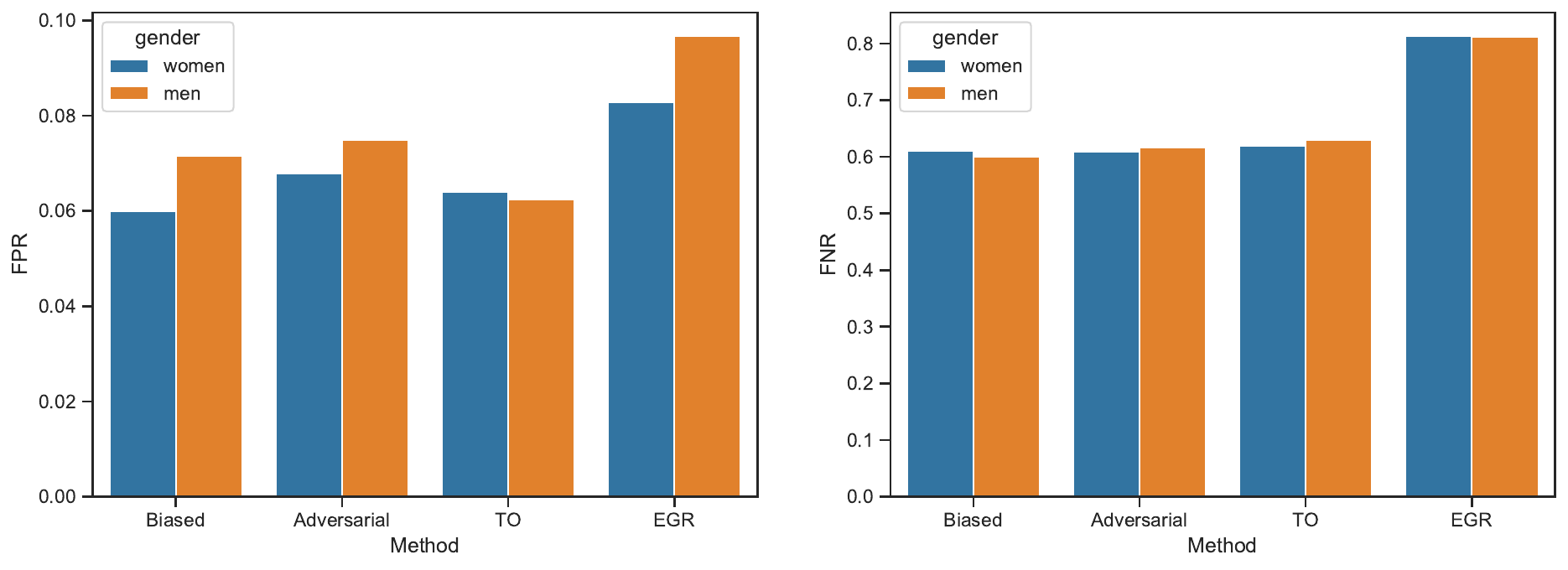}
    \caption{Final decision rates on Equilized Odds (D3): FPR (left) and FNR (right) values achieved by bias mitigation models for each sensitive group Adult (top) and Credit (bottom) datasets.}
    \label{fig:q3-FPR}
\end{figure}



To put the previous observations into perspective, we also look at the values of $\mathbb{E}(\hat{Y})$ (for DP methods) and False Positive and False Negative Rates (for EO methods) for the fair models, for each sensitive group.
Results are shown in Figure~\ref{fig:q3-EY} for the DP methods, and in Figure ~\ref{fig:q3-FPR} for EO methods, both for the Adult and Credit datasets. These results highlight additional differences between bias mitigation approaches. Indeed, despite achieving similarly good fairness and accuracy scores (cf. Table~\ref{tab:metrics}), i.e. comparable rates of population accepted into the desirable class $\mathbb{E}(\hat{Y})$ for men and women (DP), the values of these rates differ significantly depending on the approach considered (e.g. for Adult, $\approx0.17$ for LFR, against $\approx0.09$ for ROC). When enforcing Equalized Odds, larger differences between methods can be observed for the False Positive Rate (e.g. for Adult, $\approx0.13$ for EGR, against $\approx0.07$ for Adversarial) than for the False Negative Rate.
Overall, these differences are allowed to exist despite these models reaching similar predictive and fairness performance due to the differences in the size of each subgroup. 
This illustrates how even for a fixed definition of fairness (i.e. the same criterion is optimized), and when similar scores are achieved, the models may still achieve fairness in different ways. In the case of ROC for instance,
fairness is achieved by decreasing the acceptation rate for men, to the point that the ratios are equivalent across subgroups, while LFR takes a more balanced approach.

This results in some models being comparable in performance, but one (e.g. ROC here) making it much more difficult for an instance to be in the positive class than a competitor (e.g. LFR). Although rarely discussed, these differences are crucial, and even more so in the case of models being used to make critical decisions, such as in the context of personal finance or insurance. Practically, this could pose problems in situations in which there is a minimum or a cap on the percentage of people in the desirable class - for instance, if a hiring algorithm were debiased, the number of positions is fixed and all methods would therefore not be suited.


\subsection{D4: Affected subpopulations}

\begin{figure*}[t]
    \centering
    \includegraphics[width=0.45\linewidth]{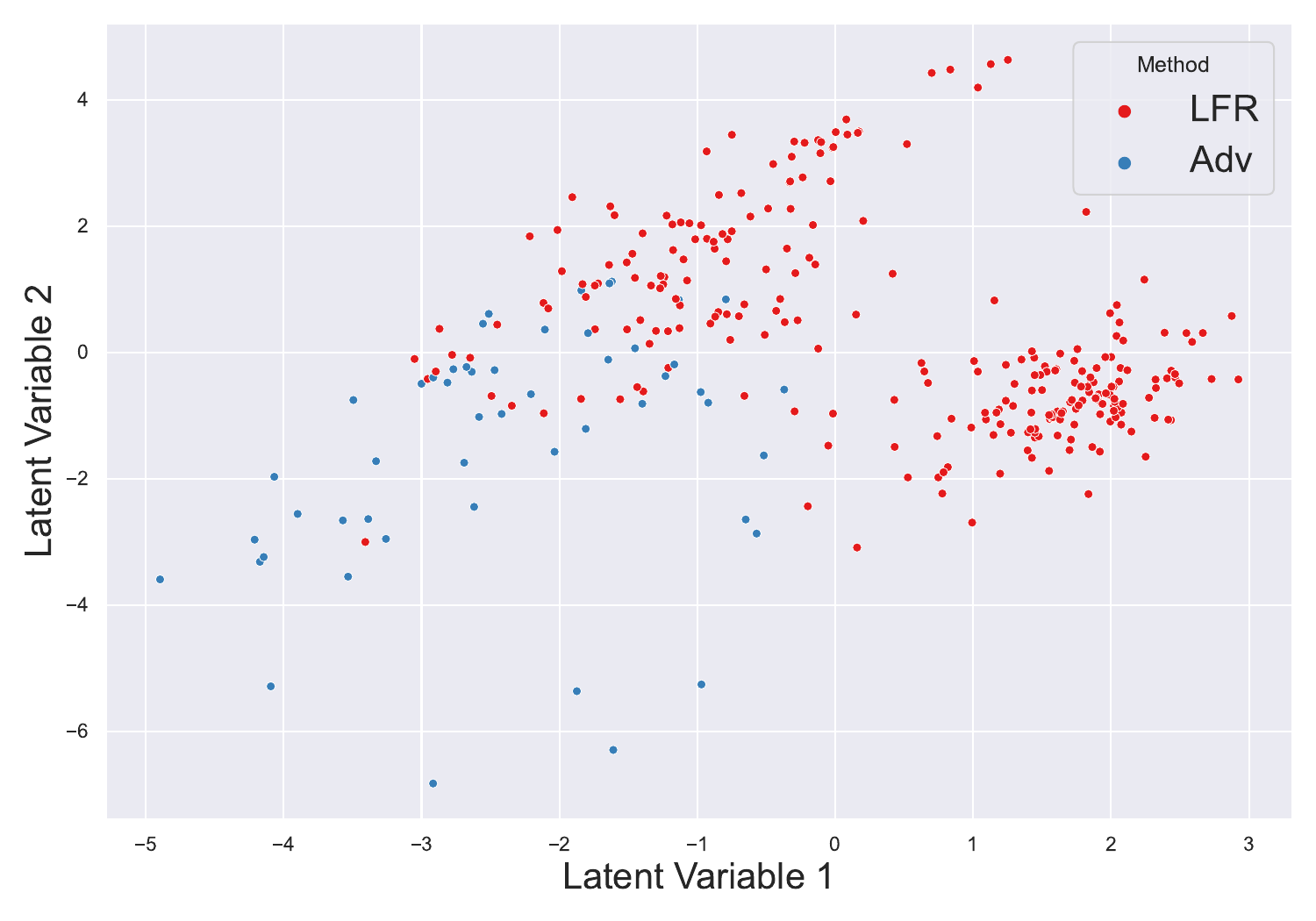}
    \includegraphics[width=0.45\linewidth]{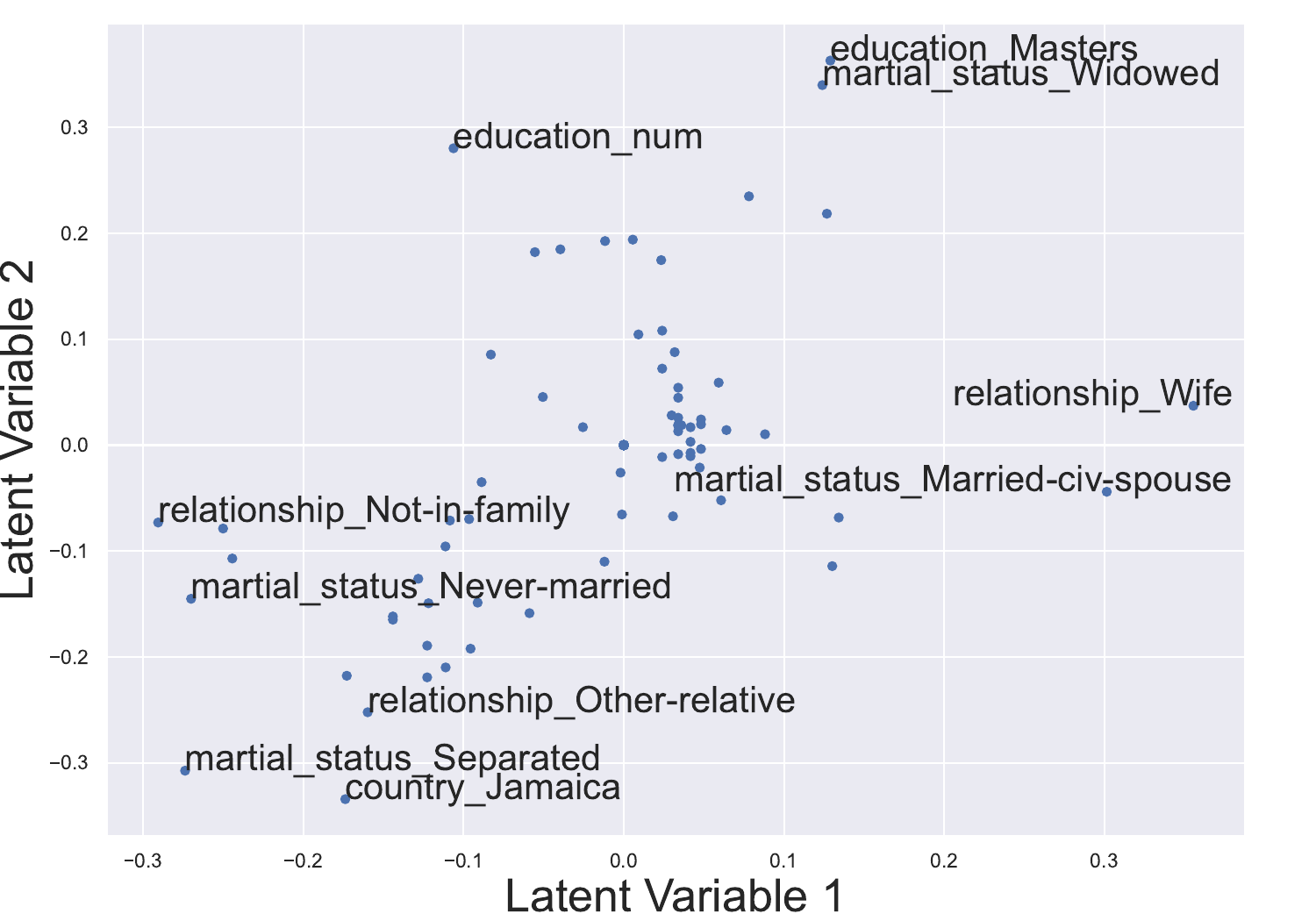}
    \includegraphics[width=0.45\linewidth]{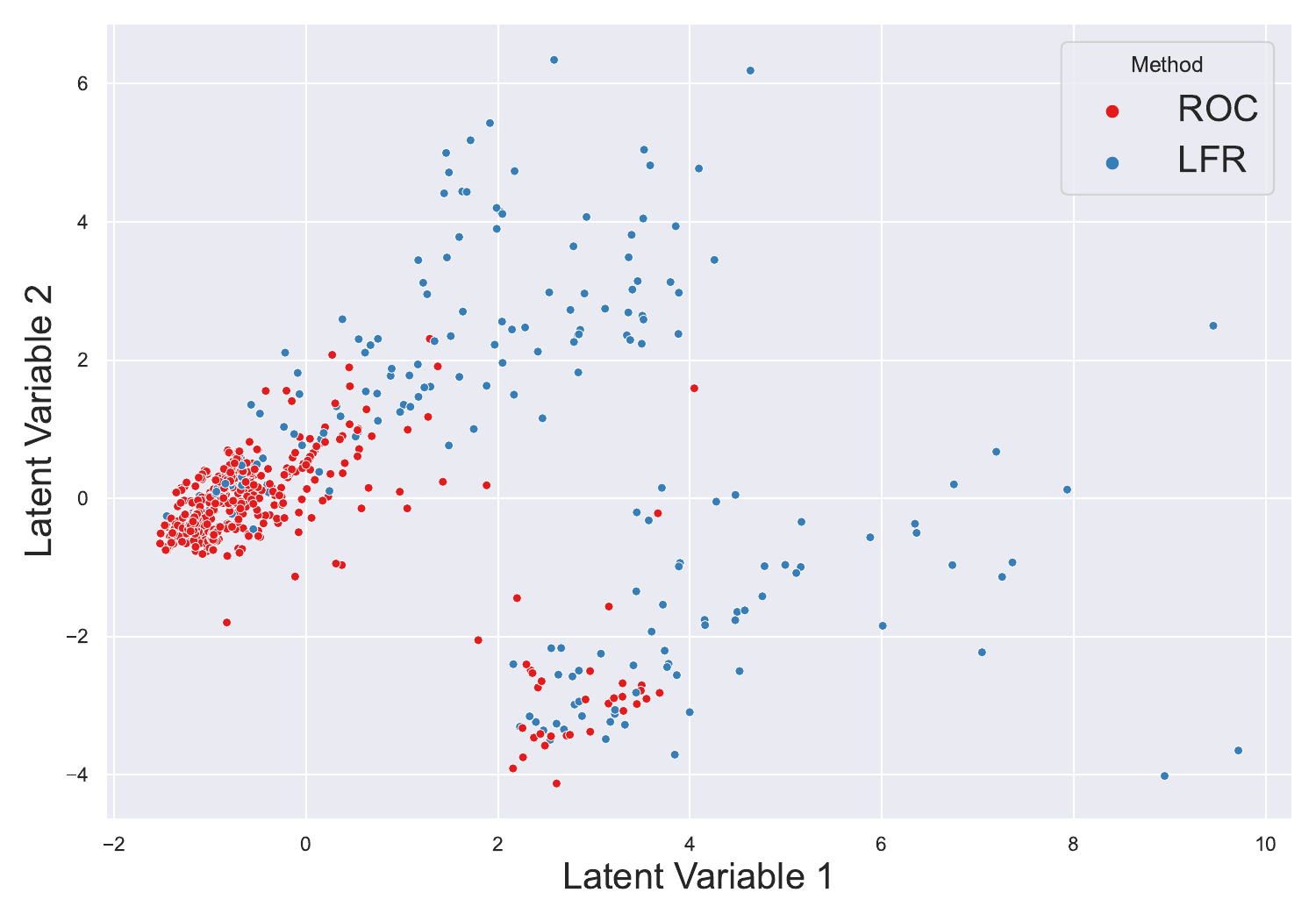}
    \includegraphics[width=0.45\linewidth]{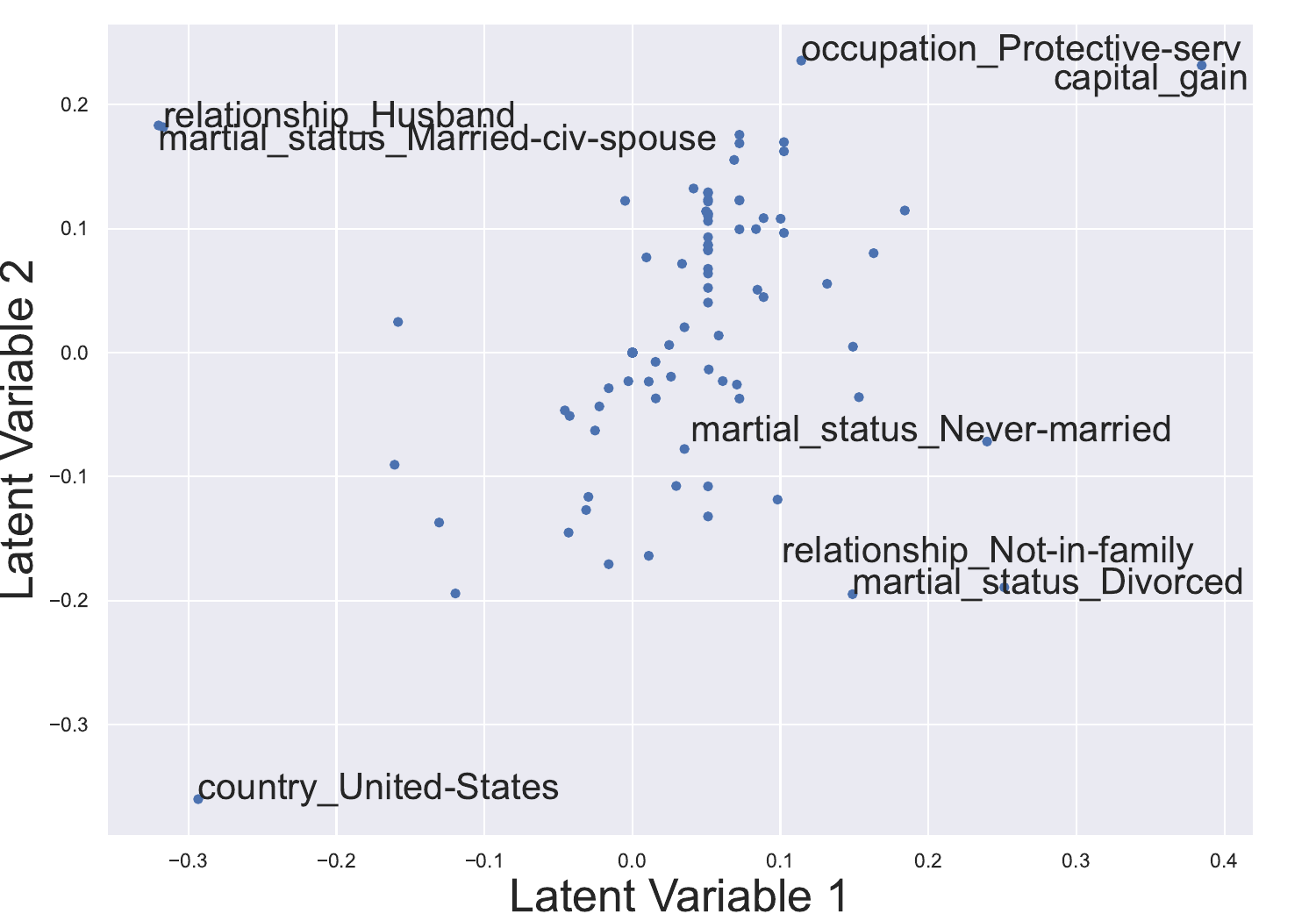}
    \caption{Results of two PLS-DA applied to the women targeted by LFR and Adversarial (top row) and men targeted by LFR and ROC (bottom row) for the Adult Dataset.}
    \label{fig:q3_2_plsda}
\end{figure*}

We push further the characterization of the populations targeted by each debiasing method by investigating and characterizing their distributions in the feature space. As described in the methodology, we conduct a Partial Least Squares Discriminant Analysis (PLS-DA) over the individuals of the Adult dataset whose predictions were changed by the fairness algorithms. This allows a 2-by-2 comparison of methods, that we conduct separately for men and women to allow a better understanding of the underlying mechanisms.
To focus the analysis on the differences between methods, we exclude the individuals whose prediction was changed by both methods (i.e., for two models $g_1$ and $g_2$ we focus on the subset $\{x \in (\Delta_{g_1}\cup\Delta_{g_2})  \setminus (\Delta_{g_1}\cap\Delta_{g_2})\}$).
Figure~\ref{fig:q3_2_plsda} shows the results of this analysis applied to women that were either targeted by LFR (red) or by Adversarial (blue) 
(top row) and on men being targeted by LFR (blue) and ROC (red) (bottom row).   
Given, for each row, the two subpopulations of individuals affected in the debiasing by the aforementioned methods, PLS-DA aims to learns a subspace that allows to linearly separate these subpopulations: the projection of these populations on the two most important components describing this subspace can be seen in the left column of Figure~\ref{fig:q3_2_plsda}, illustrating their discriminant power. The final AUC scores for each linear model are respectively $0.95$ and $0.91$

In the right column, the correlation matrix between the features describing the input space~$X$ and the latent variables learned by PLS-DA is shown: the closer to the center $(0., 0.)$ the features are, and the less they contribute to the two components. On the contrary, the further they are, the higher their importance is in constructing the latent variables, and thus in separating the two subpopulations. 
In the top row, we observe that the two algorithms thus seem to focus on women of different marital and relationship statuses and education.
For men on the other hand, the origin country (LFR rather targeting non-American people compared to ROC) and the capital gain (LFR targeting higher values) seem to be the most important features. These qualitative results highlight the diversity in the strategies employed by the considered methods: besides targeting different instances, they thus also focus on different regions of the feature space. 

When retraining the bias mitigation algorithms and replicating this same experiment, we do not observe any consistency in the features selected by each method. This aligns with the results of in Table~\ref{tab:stability-dp}): retraining the same algorithms may lead to drastically different models, and no conclusion can be drawn on specific strategies employed by the considered methods. 
Yet, the differences highlighted underline the -generally hidden- diversity of the fairness methods, and raise questions about their practical use.

\subsection{D5: Neglected Subpopulations}
\label{sec:question4}


\begin{figure}[t]
    \centering
    \includegraphics[width=0.8\linewidth]{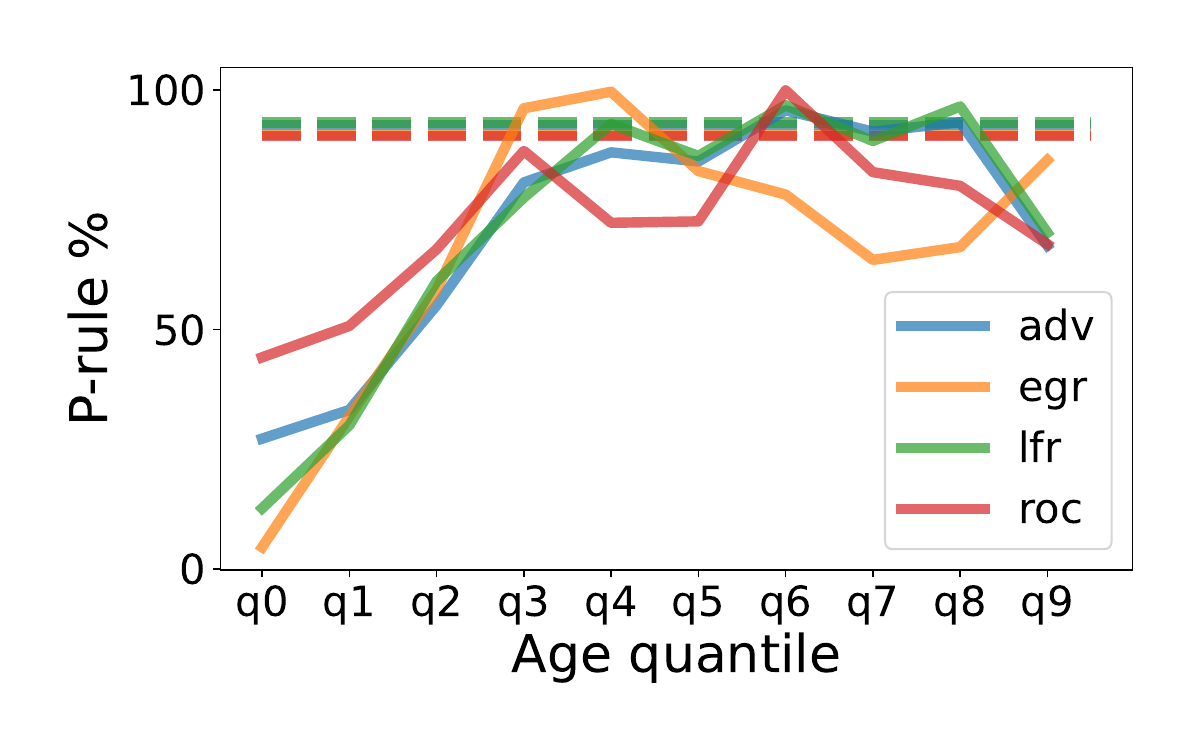}
    \includegraphics[width=0.8\linewidth]{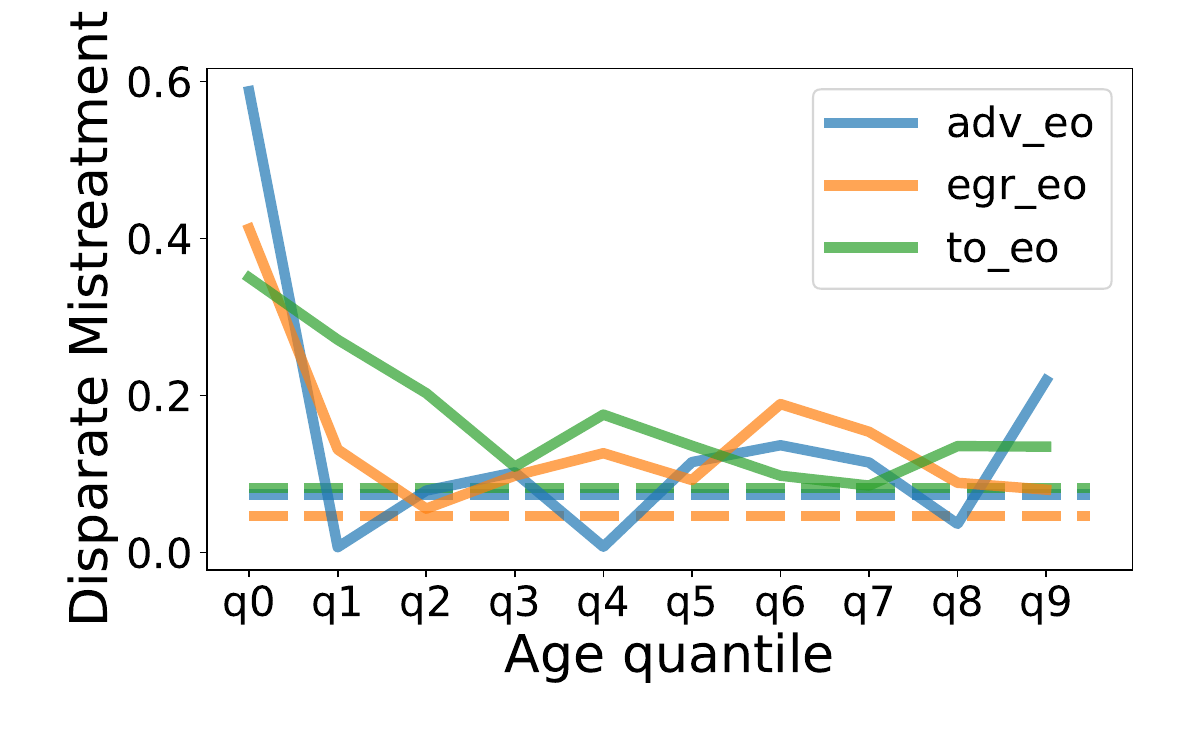}
    \caption{Global (dashed lines) and Local (full lines) fairness scores for DP (top) and EO (bottom) methods on Adult (D5).}
    \label{fig:local-bias}
\end{figure}

This section investigates the local fairness—or fairness within specific subpopulations in the data. For this purpose, we replicate the experiment of~\cite{grari2023fairness} and segment the datasets along features of~$X$. These subgroups should be defined by the practitioner depending on the context; for the Adult dataset, we segment the data along the feature $Age$ in 10 quantiles.
We then measure, in these segments, the local fairness scores, given by the $\%$p-rule for DP and DM for EO. The local (full lines) and global (dashed lines) fairness scores are shown in Figure~\ref{fig:local-bias}, for the DP (top) and EO (bottom) methods. 

Despite all models achieving good fairness scores overall, these results do not hold for all segments of the population. In particular in the first two deciles (corresponding to individuals younger than 26 years-old), the $\%$p-rule scores are consistently below $50\%$. An exception is ROC, which achieves slightly better results. 
One interpretation for these behaviors is that in these segments, the distributions of the sensitive and the target variables are the most different imbalanced. As a result, the various bias mitigation algorithms seemingly fail to enforce fairness for these subpopulations. 
Several approaches have been proposed to ensure better fairness generalization~\cite{ferry2023improving,grari2023fairness}, and could therefore be considered to mitigate this issue.



\section{Discussion and Perspectives}
\label{sec:conclusion}


The results of the study presented in this paper exhibit strong differences between bias mitigation approaches, along various dimensions: the size of the impact of the debiasing process, how the individuals are impacted by it, which individuals are impacted, and which have been, on the contrary, neglected by these methods.
Additionally, some of these approaches have been shown to provide little stability in terms of targeted populations when retrained. 
The panel of popular group fairness methods tested in this paper underline the general aspect of the obtained results and the conclusions we are able to draw from them. This study thus confirms the importance of accounting for model multiplicity in the debiasing process, and the harmful arbitrariness that results from failing to do so. It also allows us to identify promising research directions, that we discuss below.


\paragraph{Perspective 1: Beyond Demographic Parity and Equalized Odds.}
The behaviors highlighted in the study raise the question of the extent to which these observations generalize to other algorithmic fairness contexts. Thus, it would be interesting to conduct a similar study to other group fairness criteria, such as \emph{Well calibration}~\cite{kleinberg2016inherent}, and assess whether the constraints imposed by these more specified criteria still leave room for important predictive multiplicity. 
Furthermore, addressing similar questions to contexts beyond group fairness is also a promising research direction, as the behaviors observed have mainly been attributed to the global aspect of the considered criteria, as discussed by~\citet{binns2020apparent}. In particular, addressing multiplicity in Counterfactual Fairness~\cite{kusner2017counterfactual}, which aims to guarantee the statistical parity on the outcomes generated from causal intervention, as measured for instance by the \emph{Total Causal Effect} criterion, is a promising research direction.

\begin{figure}
    \centering
    \includegraphics[width=0.5\linewidth]{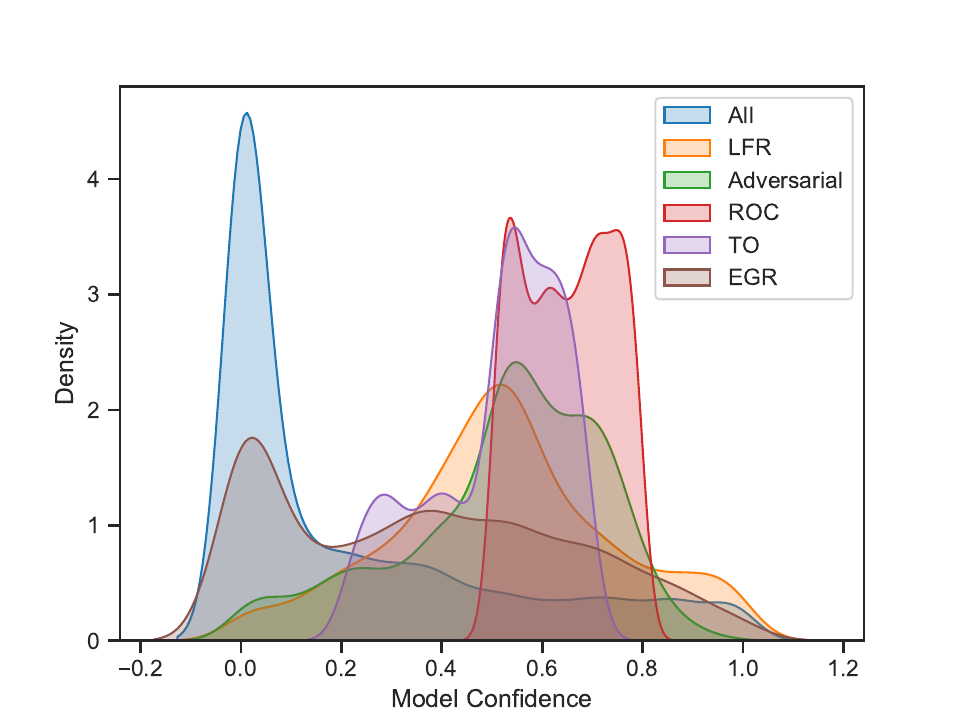}
    \includegraphics[width=0.43\linewidth]{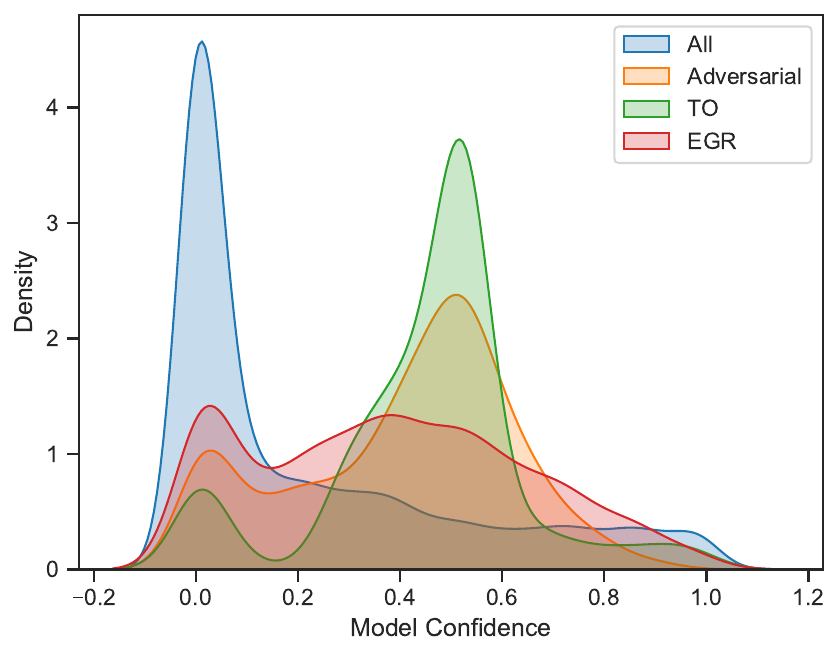}
    \caption{Biased model's confidence distribution, for all instances of the test set ("All", in blue) and for the sets $\Delta_g$ for fair models $g$ described in the legend. Left is DP, right is EO.}
    \label{fig:jml}
\end{figure}

\paragraph{Perspective 2: Further characterization of the debiasing process.}
Additionally, the question can be raised of whether our framework still misses interesting dimensions to better understand debiasing and its impacts. For instance, besides understanding \emph{who} is targeted and \emph{how}, as explored in this study, the crucial question of \emph{why} some individuals are targeted remains unanswered. 
To formulate a preliminary answer to this question, we propose in Figure~\ref{fig:jml} an exploratory experiment on this topic, for both Demographic Parity (left) and Equalized Odds (right). The distribution of the biased model confidence (x-axis) for instances of $X_{val}$ ("All", in blue), and for instances belonging to each subset $\Delta_{g_i}$. We observe that most debiasing methods target instances for which the model is the least confident (confidence close to $0.5$), suggesting that individuals closer to the decision boundary tend to be targeted more easily. Nevertheless, further work is required to rigorously answer this question.
On a different note, aiming to characterize the debiasing process on non-quantitative dimensions, such as assessing the legal, social or business implications of implementing one model rather than another, remains a question of utmost importance, that we hope can be more easily addressed by leveraging FRAME.

\paragraph{Perspective 3: Towards better fair models.}
Another possible way forward for the field of Fair ML would be working towards removing the source of arbitrariness. This could be potentially achieved by aiming for more specification for the task of bias mitigation. Beyond questioning the relevance of current group fairness criteria, a possibility would thus be to design fairness algorithms that would select which populations should get their predictions altered positively or negatively in less arbitrary manners. In that regard, we believe each dimension presented in this paper raises questions that deserve further research, e.g. working on the calibration of DP and EO models (Dimension 3), or on designing more change-efficient models (Dimension 1).
Another example, raised by Dimension 4, would be to rely on user- or society-defined secondary criteria to define some prioritization of individuals that may be affected by the debiasing process. 
In the meantime, having more transparent debiasing processes (i.e. not only transparent fair models, but understanding the impact of bias mitigation) is of utmost importance.



\section{Acknowledgements}
Jean-Michel Loubes’ research is partially supported by the AI  Interdisciplinary Institute ANITI.

\bibliography{biblio}

\begin{thebibliography}{59}
\providecommand{\natexlab}[1]{#1}

\bibitem[{Adel et~al.(2019)Adel, Valera, Ghahramani, and Weller}]{adel2019one}
Adel, T.; Valera, I.; Ghahramani, Z.; and Weller, A. 2019.
\newblock One-network adversarial fairness.
\newblock In \emph{Proceedings of the AAAI Conference on Artificial Intelligence}, volume~33, 2412--2420.

\bibitem[{Agarwal et~al.(2018)Agarwal, Beygelzimer, Dud{\'\i}k, Langford, and Wallach}]{agarwal2018reductions}
Agarwal, A.; Beygelzimer, A.; Dud{\'\i}k, M.; Langford, J.; and Wallach, H. 2018.
\newblock A reductions approach to fair classification.
\newblock In \emph{International Conference on Machine Learning}, 60--69. PMLR.

\bibitem[{A{\"\i}vodji et~al.(2019)A{\"\i}vodji, Ferry, Gambs, Huguet, and Siala}]{aivodji2019learning}
A{\"\i}vodji, U.; Ferry, J.; Gambs, S.; Huguet, M.-J.; and Siala, M. 2019.
\newblock Learning fair rule lists.
\newblock \emph{arXiv preprint arXiv:1909.03977}.

\bibitem[{Angwin et~al.(2016)Angwin, Larson, Mattu, and Kirchner}]{angwin2016machine}
Angwin, J.; Larson, J.; Mattu, S.; and Kirchner, L. 2016.
\newblock {Machine bias. ProPublica, May 23, 2016}.

\bibitem[{Balayn et~al.(2023)Balayn, Yurrita, Yang, and Gadiraju}]{balayn2023fairness}
Balayn, A.; Yurrita, M.; Yang, J.; and Gadiraju, U. 2023.
\newblock “eazFairness Toolkits, A Checkbox Culture?” On the Factors that Fragment Developer Practices in Handling Algorithmic Harms.
\newblock In \emph{Proceedings of the 2023 AAAI/ACM Conference on AI, Ethics, and Society}, 482--495.

\bibitem[{Becker and Kohavi(1996)}]{misc_adult_2}
Becker, B.; and Kohavi, R. 1996.
\newblock {Adult}.
\newblock UCI Machine Learning Repository.

\bibitem[{Bellamy et~al.(2018)Bellamy, Dey, Hind, Hoffman, Houde, Kannan, Lohia, Martino, Mehta, Mojsilovic, Nagar, Ramamurthy, Richards, Saha, Sattigeri, Singh, Varshney, and Zhang}]{aif360}
Bellamy, R. K.~E.; Dey, K.; Hind, M.; Hoffman, S.~C.; Houde, S.; Kannan, K.; Lohia, P.; Martino, J.; Mehta, S.; Mojsilovic, A.; Nagar, S.; Ramamurthy, K.~N.; Richards, J.; Saha, D.; Sattigeri, P.; Singh, M.; Varshney, K.~R.; and Zhang, Y. 2018.
\newblock {AI Fairness} 360: An Extensible Toolkit for Detecting, Understanding, and Mitigating Unwanted Algorithmic Bias.
\newblock \emph{arXiv:1810.01943}.

\bibitem[{Besse et~al.(2022)Besse, del Barrio, Gordaliza, Loubes, and Risser}]{besse2022survey}
Besse, P.; del Barrio, E.; Gordaliza, P.; Loubes, J.-M.; and Risser, L. 2022.
\newblock A survey of bias in machine learning through the prism of statistical parity.
\newblock \emph{The American Statistician}, 76(2): 188--198.

\bibitem[{Binns(2020)}]{binns2020apparent}
Binns, R. 2020.
\newblock On the apparent conflict between individual and group fairness.
\newblock In \emph{Proceedings of the 2020 conference on fairness, accountability, and transparency}, 514--524.

\bibitem[{Bird et~al.(2020)Bird, Dud{\'i}k, Edgar, Horn, Lutz, Milan, Sameki, Wallach, and Walker}]{fairlearn}
Bird, S.; Dud{\'i}k, M.; Edgar, R.; Horn, B.; Lutz, R.; Milan, V.; Sameki, M.; Wallach, H.; and Walker, K. 2020.
\newblock Fairlearn: A toolkit for assessing and improving fairness in {AI}.
\newblock Technical report, Microsoft.

\bibitem[{Black, Raghavan, and Barocas(2022)}]{black2022model}
Black, E.; Raghavan, M.; and Barocas, S. 2022.
\newblock Model multiplicity: Opportunities, concerns, and solutions.
\newblock In \emph{2022 ACM Conference on Fairness, Accountability, and Transparency}, 850--863.

\bibitem[{Burgeno and Joslyn(2020)}]{burgeno2020impact}
Burgeno, J.~N.; and Joslyn, S.~L. 2020.
\newblock The impact of weather forecast inconsistency on user trust.
\newblock \emph{Weather, climate, and society}, 12(4): 679--694.

\bibitem[{Carvalho, Pereira, and Cardoso(2019)}]{carvalho2019machine}
Carvalho, D.~V.; Pereira, E.~M.; and Cardoso, J.~S. 2019.
\newblock Machine learning interpretability: A survey on methods and metrics.
\newblock \emph{Electronics}, 8(8): 832.

\bibitem[{Creel and Hellman(2022)}]{creel2022algorithmic}
Creel, K.; and Hellman, D. 2022.
\newblock The algorithmic leviathan: Arbitrariness, fairness, and opportunity in algorithmic decision-making systems.
\newblock \emph{Canadian Journal of Philosophy}, 52(1): 26--43.

\bibitem[{D'Amour et~al.(2022)D'Amour, Heller, Moldovan, Adlam, Alipanahi, Beutel, Chen, Deaton, Eisenstein, Hoffman et~al.}]{damour2022underspecification}
D'Amour, A.; Heller, K.; Moldovan, D.; Adlam, B.; Alipanahi, B.; Beutel, A.; Chen, C.; Deaton, J.; Eisenstein, J.; Hoffman, M.~D.; et~al. 2022.
\newblock Underspecification presents challenges for credibility in modern machine learning.
\newblock \emph{Journal of Machine Learning Research}, 23(226): 1--61.

\bibitem[{Del~Barrio, Gordaliza, and Loubes(2020)}]{del2020review}
Del~Barrio, E.; Gordaliza, P.; and Loubes, J.-M. 2020.
\newblock Review of mathematical frameworks for fairness in machine learning.
\newblock \emph{arXiv preprint arXiv:2005.13755}.

\bibitem[{Dwork et~al.(2012)Dwork, Hardt, Pitassi, Reingold, and Zemel}]{dwork2012fairness}
Dwork, C.; Hardt, M.; Pitassi, T.; Reingold, O.; and Zemel, R. 2012.
\newblock Fairness through awareness.
\newblock In \emph{Proceedings of the 3rd innovations in theoretical computer science conference}, 214--226.

\bibitem[{D’Amour et~al.(2020)D’Amour, Heller, Moldovan, Adlam, Alipanahi, Beutel, Chen, Deaton, Eisenstein, Hoffman et~al.}]{d2020underspecification}
D’Amour, A.; Heller, K.; Moldovan, D.; Adlam, B.; Alipanahi, B.; Beutel, A.; Chen, C.; Deaton, J.; Eisenstein, J.; Hoffman, M.~D.; et~al. 2020.
\newblock Underspecification presents challenges for credibility in modern machine learning.
\newblock \emph{Journal of Machine Learning Research}.

\bibitem[{Ferry et~al.(2023)Ferry, Aivodji, Gambs, Huguet, and Siala}]{ferry2023improving}
Ferry, J.; Aivodji, U.; Gambs, S.; Huguet, M.-J.; and Siala, M. 2023.
\newblock Improving fairness generalization through a sample-robust optimization method.
\newblock \emph{Machine Learning}, 112(6): 2131--2192.

\bibitem[{Friedler et~al.(2019)Friedler, Scheidegger, Venkatasubramanian, Choudhary, Hamilton, and Roth}]{friedler2019comparative}
Friedler, S.~A.; Scheidegger, C.; Venkatasubramanian, S.; Choudhary, S.; Hamilton, E.~P.; and Roth, D. 2019.
\newblock A comparative study of fairness-enhancing interventions in machine learning.
\newblock In \emph{Proceedings of the conference on fairness, accountability, and transparency}, 329--338.

\bibitem[{Goethals, Calders, and Martens(2024)}]{goethals2024beyond}
Goethals, S.; Calders, T.; and Martens, D. 2024.
\newblock Beyond Accuracy-Fairness: Stop evaluating bias mitigation methods solely on between-group metrics.
\newblock \emph{arXiv preprint arXiv:2401.13391}.

\bibitem[{Gordaliza et~al.(2019)Gordaliza, Del~Barrio, Fabrice, and Loubes}]{gordaliza2019obtaining}
Gordaliza, P.; Del~Barrio, E.; Fabrice, G.; and Loubes, J.-M. 2019.
\newblock Obtaining fairness using optimal transport theory.
\newblock In \emph{International Conference on Machine Learning}, 2357--2365.

\bibitem[{Grari et~al.(2021)Grari, Hajouji, Lamprier, and Detyniecki}]{grari2021learning}
Grari, V.; Hajouji, O.~E.; Lamprier, S.; and Detyniecki, M. 2021.
\newblock Learning Unbiased Representations via R{\'e}nyi Minimization.
\newblock In \emph{Joint European Conference on Machine Learning and Knowledge Discovery in Databases}, 749--764.

\bibitem[{Grari et~al.(2024)Grari, Laugel, Hashimoto, Detyniecki et~al.}]{grari2023fairness}
Grari, V.; Laugel, T.; Hashimoto, T.; Detyniecki, M.; et~al. 2024.
\newblock On the Fairness ROAD: Robust Optimization for Adversarial Debiasing.
\newblock In \emph{The Twelfth International Conference on Learning Representations}.

\bibitem[{Grari et~al.(2019)Grari, Ruf, Lamprier, and Detyniecki}]{grari2019fair}
Grari, V.; Ruf, B.; Lamprier, S.; and Detyniecki, M. 2019.
\newblock Fair adversarial gradient tree boosting.
\newblock In \emph{2019 IEEE International Conference on Data Mining (ICDM)}, 1060--1065. IEEE.

\bibitem[{Hardt, Price, and Srebro(2016)}]{hardt2016equality}
Hardt, M.; Price, E.; and Srebro, N. 2016.
\newblock Equality of opportunity in supervised learning.
\newblock \emph{Advances in neural information processing systems}, 29.

\bibitem[{Hashimoto et~al.(2018)Hashimoto, Srivastava, Namkoong, and Liang}]{hashimoto2018fairness}
Hashimoto, T.; Srivastava, M.; Namkoong, H.; and Liang, P. 2018.
\newblock Fairness without demographics in repeated loss minimization.
\newblock In \emph{International Conference on Machine Learning}, 1929--1938. PMLR.

\bibitem[{Hort et~al.(2022)Hort, Chen, Zhang, Sarro, and Harman}]{hort2022survey}
Hort, M.; Chen, Z.; Zhang, J.~M.; Sarro, F.; and Harman, M. 2022.
\newblock Bias mitigation for machine learning classifiers: A comprehensive survey.
\newblock \emph{arXiv:2207.07068}.

\bibitem[{Jobin, Ienca, and Vayena(2019)}]{jobin2019global}
Jobin, A.; Ienca, M.; and Vayena, E. 2019.
\newblock The global landscape of AI ethics guidelines.
\newblock \emph{Nature Machine Intelligence}, 1(9): 389--399.

\bibitem[{Kamiran and Calders(2009)}]{kamiran2009classifying}
Kamiran, F.; and Calders, T. 2009.
\newblock Classifying without discriminating.
\newblock In \emph{2009 2nd international conference on computer, control and communication}, 1--6.

\bibitem[{Kamiran, Karim, and Zhang(2012)}]{kamiran2012roc}
Kamiran, F.; Karim, A.; and Zhang, X. 2012.
\newblock Decision Theory for Discrimination-Aware Classification.
\newblock In \emph{2012 IEEE 12th International Conference on Data Mining}, 924--929.

\bibitem[{Kamishima et~al.(2012)Kamishima, Akaho, Asoh, and Sakuma}]{kamishma2012prejudice}
Kamishima, T.; Akaho, S.; Asoh, H.; and Sakuma, J. 2012.
\newblock Fairness-Aware Classifier with Prejudice Remover Regularizer.
\newblock In Flach, P.~A.; De~Bie, T.; and Cristianini, N., eds., \emph{Machine Learning and Knowledge Discovery in Databases}, 35--50.

\bibitem[{Kamp, Zhao, and Kutty(2021)}]{kamp2021robustness}
Kamp, S.; Zhao, A. L.~L.; and Kutty, S. 2021.
\newblock Robustness of fairness: An experimental analysis.
\newblock In \emph{Joint European Conference on Machine Learning and Knowledge Discovery in Databases}, 591--606. Springer.

\bibitem[{Kleinberg, Mullainathan, and Raghavan(2016)}]{kleinberg2016inherent}
Kleinberg, J.; Mullainathan, S.; and Raghavan, M. 2016.
\newblock Inherent trade-offs in the fair determination of risk scores.
\newblock \emph{arXiv preprint arXiv:1609.05807}.

\bibitem[{Krishna et~al.(2022)Krishna, Han, Gu, Pombra, Jabbari, Wu, and Lakkaraju}]{krishna2022disagreement}
Krishna, S.; Han, T.; Gu, A.; Pombra, J.; Jabbari, S.; Wu, S.; and Lakkaraju, H. 2022.
\newblock The disagreement problem in explainable machine learning: A practitioner's perspective.
\newblock \emph{arXiv preprint arXiv:2202.01602}.

\bibitem[{Kusner et~al.(2017)Kusner, Loftus, Russell, and Silva}]{kusner2017counterfactual}
Kusner, M.~J.; Loftus, J.; Russell, C.; and Silva, R. 2017.
\newblock Counterfactual fairness.
\newblock \emph{Advances in neural information processing systems}, 30.

\bibitem[{Lipton, McAuley, and Chouldechova(2018)}]{lipton2018does}
Lipton, Z.; McAuley, J.; and Chouldechova, A. 2018.
\newblock Does mitigating ML's impact disparity require treatment disparity?
\newblock \emph{Advances in neural information processing systems}, 31.

\bibitem[{Long et~al.(2024)Long, Hsu, Alghamdi, and Calmon}]{long2024individual}
Long, C.; Hsu, H.; Alghamdi, W.; and Calmon, F. 2024.
\newblock Individual Arbitrariness and Group Fairness.
\newblock \emph{Advances in Neural Information Processing Systems}, 36.

\bibitem[{Lundberg and Lee(2017)}]{lundberg2017unified}
Lundberg, S.~M.; and Lee, S.-I. 2017.
\newblock A unified approach to interpreting model predictions.
\newblock \emph{Advances in neural information processing systems}, 30.

\bibitem[{Marchiori~Manerba and Guidotti(2022)}]{manerba2022investigating}
Marchiori~Manerba, M.; and Guidotti, R. 2022.
\newblock Investigating Debiasing Effects on Classification and Explainability.
\newblock In \emph{Proceedings of the 2022 AAAI/ACM Conference on AI, Ethics, and Society}, 468–478.

\bibitem[{Marx, Calmon, and Ustun(2020)}]{marx2020predictive}
Marx, C.; Calmon, F.; and Ustun, B. 2020.
\newblock Predictive multiplicity in classification.
\newblock In \emph{International Conference on Machine Learning}, 6765--6774.

\bibitem[{Mehrabi et~al.(2021)Mehrabi, Morstatter, Saxena, Lerman, and Galstyan}]{mehrabi2021survey}
Mehrabi, N.; Morstatter, F.; Saxena, N.; Lerman, K.; and Galstyan, A. 2021.
\newblock A Survey on Bias and Fairness in Machine Learning.
\newblock \emph{ACM Comput. Surv.}, 54(6).

\bibitem[{Mittelstadt, Wachter, and Russell(2023)}]{mittelstadt2023unfairness}
Mittelstadt, B.; Wachter, S.; and Russell, C. 2023.
\newblock The Unfairness of Fair Machine Learning: Levelling down and strict egalitarianism by default.
\newblock \emph{arXiv preprint arXiv:2302.02404}.

\bibitem[{Moro, Cortez, and Rita(2014)}]{bankdataset}
Moro, S.; Cortez, P.; and Rita, P. 2014.
\newblock A data-driven approach to predict the success of bank telemarketing.
\newblock \emph{Decision Support Systems}, 62: 22--31.

\bibitem[{Nanda et~al.(2021)Nanda, Dooley, Singla, Feizi, and Dickerson}]{nanda2021fairness}
Nanda, V.; Dooley, S.; Singla, S.; Feizi, S.; and Dickerson, J.~P. 2021.
\newblock Fairness through robustness: Investigating robustness disparity in deep learning.
\newblock In \emph{Proceedings of the 2021 ACM Conference on Fairness, Accountability, and Transparency}, 466--477.

\bibitem[{Nordholt(2005)}]{dutchdataset}
Nordholt, E.~S. 2005.
\newblock The Dutch virtual Census 2001: A new approach by combining different sources.
\newblock \emph{Statistical Journal of the United Nations Economic Commission for Europe}, 22(1): 25--37.

\bibitem[{Quy et~al.(2022)Quy, Roy, Iosifidis, Zhang, and Ntoutsi}]{lequy202survey}
Quy, T.~L.; Roy, A.; Iosifidis, V.; Zhang, W.; and Ntoutsi, E. 2022.
\newblock A survey on datasets for fairness-aware machine learning.
\newblock \emph{{WIREs} Data Mining and Knowledge Discovery}, 12(3).

\bibitem[{Renard, Laugel, and Detyniecki(2021)}]{renard2021understanding}
Renard, X.; Laugel, T.; and Detyniecki, M. 2021.
\newblock Understanding Prediction Discrepancies in Machine Learning Classifiers.
\newblock \emph{arXiv preprint arXiv:2104.05467}.

\bibitem[{Romei and Ruggieri(2013)}]{Romei2013AMS}
Romei, A.; and Ruggieri, S. 2013.
\newblock A multidisciplinary survey on discrimination analysis.
\newblock \emph{The Knowledge Engineering Review}, 29: 582 -- 638.

\bibitem[{Selbst et~al.(2019)Selbst, Boyd, Friedler, Venkatasubramanian, and Vertesi}]{selbst2019fairness}
Selbst, A.~D.; Boyd, D.; Friedler, S.~A.; Venkatasubramanian, S.; and Vertesi, J. 2019.
\newblock Fairness and abstraction in sociotechnical systems.
\newblock In \emph{Proceedings of the conference on fairness, accountability, and transparency}, 59--68.

\bibitem[{Wachter, Mittelstadt, and Russell(2021{\natexlab{a}})}]{wachter2021why}
Wachter, S.; Mittelstadt, B.; and Russell, C. 2021{\natexlab{a}}.
\newblock Why fairness cannot be automated: Bridging the gap between EU non-discrimination law and AI.
\newblock \emph{Computer Law \& Security Review}, 41: 105567.

\bibitem[{Wachter, Mittelstadt, and Russell(2021{\natexlab{b}})}]{wachter2021fairness}
Wachter, S.; Mittelstadt, B.; and Russell, C. 2021{\natexlab{b}}.
\newblock Why fairness cannot be automated: Bridging the gap between EU non-discrimination law and AI.
\newblock \emph{Computer Law \& Security Review}, 41: 105567.

\bibitem[{Wold, Sj{\"o}str{\"o}m, and Eriksson(2001)}]{wold2001pls}
Wold, S.; Sj{\"o}str{\"o}m, M.; and Eriksson, L. 2001.
\newblock PLS-regression: a basic tool of chemometrics.
\newblock \emph{Chemometrics and intelligent laboratory systems}, 58(2): 109--130.

\bibitem[{Xu et~al.(2018)Xu, Yuan, Zhang, and Wu}]{xu2018fairgan}
Xu, D.; Yuan, S.; Zhang, L.; and Wu, X. 2018.
\newblock FairGAN: Fairness-aware Generative Adversarial Networks.
\newblock In \emph{2018 IEEE International Conference on Big Data (Big Data)}, 570--575.

\bibitem[{Yeh and Lien(2009)}]{creditdataset}
Yeh, I.-C.; and Lien, C.-h. 2009.
\newblock The comparisons of data mining techniques for the predictive accuracy of probability of default of credit card clients.
\newblock \emph{Expert systems with applications}, 36(2): 2473--2480.

\bibitem[{Zafar et~al.(2017)Zafar, Valera, Gomez~Rodriguez, and Gummadi}]{zafar2017fairness}
Zafar, M.~B.; Valera, I.; Gomez~Rodriguez, M.; and Gummadi, K.~P. 2017.
\newblock Fairness beyond disparate treatment \& disparate impact: Learning classification without disparate mistreatment.
\newblock In \emph{Proceedings of the 26th international conference on world wide web}, 1171--1180.

\bibitem[{Zemel et~al.(2013)Zemel, Wu, Swersky, Pitassi, and Dwork}]{zemel2013learning}
Zemel, R.; Wu, Y.; Swersky, K.; Pitassi, T.; and Dwork, C. 2013.
\newblock Learning fair representations.
\newblock In \emph{International conference on machine learning}, 325--333.

\bibitem[{Zhang, Lemoine, and Mitchell(2018)}]{zhang2018adversarial}
Zhang, B.~H.; Lemoine, B.; and Mitchell, M. 2018.
\newblock Mitigating Unwanted Biases with Adversarial Learning.

\bibitem[{{\v{Z}}liobaite, Kamiran, and Calders(2011)}]{vzliobaite2011handling}
{\v{Z}}liobaite, I.; Kamiran, F.; and Calders, T. 2011.
\newblock Handling conditional discrimination.
\newblock In \emph{2011 IEEE 11th international conference on data mining}, 992--1001. IEEE.

\end{thebibliography}

\newpage
\section{Appendix}

\subsection{Datsets}

Below can be found a description of the datasets used in this study, with details on the target and sensitive variables considered.
\begin{table}[h!]
    \centering
    \begin{tabular}{|l|l|l|} \hline 
        Dataset & $S$ & $Y$ \\ \hline \hline
        \multirow{2}{*}{Adult} & \multirow{2}{*}{Gender (male)} &\multirow{2}{*}{ Income ($>50K$)} \\
        & & \\ \hline 
        \multirow{2}{*}{Dutch} & \multirow{2}{*}{Gender (male)} & \multirow{2}{8em}{Occupation Level (prestigious)} \\
        & & \\ \hline 
        \multirow{2}{*}{COMPAS} & \multirow{2}{*}{Race (white)} &\multirow{2}{7em}{Will re-offend (No)}  \\
        & & \\ \hline 
        \multirow{2}{*}{Bank} & \multirow[c]{2}{7em}{Age category ($33-60$)} & \multirow{2}{7em}{Has term deposit (Yes)} \\
        & & \\ \hline 
        \multirow{2}{*}{Credit} &\multirow{2}{*}{Gender (male)}  &\multirow{2}{7em}{Is in default payment (No)} \\
        & & \\ \hline 
    \end{tabular}
    \caption{Datasets used in analyses, with sensitive attribute $S$ and target variable $Y$. Privileged value is given for each sensitive attribute in brackets next to attribute name (i.e. Attribute name (privileged value)). Desirable outcome for $Y$ is given in the same fashion (Variable name (desirable outcome)).}
    \label{tab:datasets}
\end{table}

\subsection{Debiasing Methods}

\begin{table*}[h]
    \centering
    \begin{tabular}{|c|c|c|} \hline
        Method & Category & Metric \\ \hline \hline
        Learning Fair Representations (LFR) \cite{zemel2013learning} & Pre-processing & DP \\ \hline 
        Adversarial Debiasing (Adv$_{DP}$, Adv$_{EO}$) \cite{zhang2018adversarial} & In-processing & DP, EO \\ \hline
        Exponentiated Gradient Reduction (EGR) \cite{agarwal2018reductions} & In-processing & DP, EO \\\hline
        Reject Option Classification (ROC) \cite{kamiran2012roc} & Post-processing & DP \\ \hline
        Threshold Optimization (TO) \cite{hardt2016equality} & Post-processing & DP, EO \\\hline
    \end{tabular}
    \caption{Debiasing methods considered in our analyses, with category and metric optimized. Concerning the metrics, DP stands for Demographic Parity, and EO stands for Equalized Odds.}
    \label{tab:methods}
\end{table*}

\subsection{Implementation Details}

\subsubsection{hyperperparameters}

\begin{table*}[h!]
    \centering
    \begin{tabular}{|c|c|c|c|}
    \cline{2-4}
        \multicolumn{1}{c|}{} & Method & Library & Hyperparameters \\
        \hline
        \multirow{5}{*}{\rotatebox[origin=rB]{90}{Credit}}
        & LFR & aif360 & \{reconstruct weight: $0.01$, target weight: $1$, fairness weight: $14$ \} \\
        & Adversarial (DP) & fairlearn & \{adversary loss weight: $14$ \} \\
        & ROC & / & \{confidence threshold: $0.545$\} \\
        & Adversarial (EO) & fairlearn & $\{ \text{adversary loss weight}: 25 \}$ \\
        & Equal Odds & fairlearn & / \\
        \hline
        \hline
        
        \multirow{5}{*}{\rotatebox[origin=rB]{90}{Adult}}
        & LFR & aif360 & \{reconstruct weight: $0.01$, target weight: $10$, fairness weight: $15$ \} \\
        & Adversarial (DP) & fairlearn & \{adversary loss weight: $0.3$ \} \\
        & ROC & / & \{confidence threshold: $0.86$\} \\
        & Adversarial (EO) & fairlearn & $\{ \text{adversary loss weight}: 40 \}$ \\
        & Equal Odds & fairlearn & / \\
        \hline
        \hline

        \multirow{5}{*}{\rotatebox[origin=rB]{90}{Bank}}
        & LFR & aif360 & \{reconstruct weight: $0.01$, target weight: $1$, fairness weight: $14$ \} \\
        & Adversarial (DP) & aif360 & \{adversary loss weight: $4$ \} \\
        & ROC & / & \{confidence threshold: $0.78$\} \\
        & Adversarial (EO) & fairlearn & $\{ \text{adversary loss weight}: 2.43 \}$ \\
        & Equal Odds & fairlearn & / \\
        \hline \hline

        \multirow{5}{*}{\rotatebox[origin=rB]{90}{Dutch}}
        & LFR & aif360 & \{reconstruct weight: $0.01$, target weight: $1$, fairness weight: $8$ \} \\
        & Adversarial (DP) & fairlearn & \{adversary loss weight: $10.86$ \} \\
        & ROC & / & \{confidence threshold: $0.92$\} \\
        & Adversarial (EO) & fairlearn & $\{ \text{adversary loss weight}: 130 \}$ \\
        & Equal Odds & fairlearn & / \\
        \hline \hline

        \multirow{5}{*}{\rotatebox[origin=rB]{90}{COMPAS}}
        & LFR & aif360 & \{reconstruct weight: $0.01$, target weight: $1$, fairness weight: $2$ \} \\
        & Adversarial (DP) & aif360 & \{adversary loss weight: $3$ \} \\
        & ROC & / & \{confidence threshold: $0.81$\} \\
        & Adversarial (EO) & fairlearn & $\{ \text{adversary loss weight}: 14 \}$ \\
        & Equal Odds & fairlearn & / \\
         \hline 
    \end{tabular}
    \caption{Implementation details}
    \label{tab:hyperparams}
\end{table*}

\subsection{Performance metrics}

\begin{table*}[ht!]
    \centering
    \resizebox{0.9\textwidth}{!}
    {\begin{tabular}{|c|c|c|c||c|c|c|c|}
    \cline{2-8}
    \multicolumn{1}{c|}{} & \multicolumn{3}{c||}{Demographic Parity} & \multicolumn{4}{c|}{Equalized Odds} \\ 
    \cline{2-8} 
         \multicolumn{1}{c|}{} & Model & Accuracy [\%] & p\%-rule [\%] & Model & Accuracy [\%] & $D_{FPR}$ & $D_{TPR}$ \\ \cline{2-6} \hline
         \multirow{4}{*}{\rotatebox[origin=rB]{90}{Adult}} &
         Biased & $84.45$ & $36.02$ & Biased & $84.45$ & 6.96 & 3.78\\
         & LFR & 81.66 & $93.19$ & Adversarial & 83.42 & 1.69 & 5.73\\
         & Adversarial & 83.15 & $92.67$  & Threshold Optimizer & 83.33 & 1.16 & 7.08\\
         & ROC &  82.25  & $90.43$ & EGR & 78.64 & 3.13 & 1.53 \\  
         & EGR & 77.02 & 90.6 &&&& \\
         \hline \hline

         \multirow{4}{*}{\rotatebox[origin=rB]{90}{Bank}} &
         Biased & 89.78 & 55.48 & Biased & 89.78 & 2.5 & 6.2 \\
         & LFR & 90.6 & 81.03 & Adversarial & 89.63 & 1.87 & 1.92 \\
         & Adversarial & 89.27 & 83.95 & Threshold Optimizer & 88.92 & 3.43 &   5.36 \\
         & ROC & 89.62 & 82.68 & EGR & 88.25 & 2.47& 1.05 \\
         & EGR & 87.76 & 72.53 &&&& \\ \hline \hline

         \multirow{4}{*}{\rotatebox[origin=rB]{90}{COMPAS}} &
         Biased & 73.92 & 53.66 &  Biased & 73.92 & 24.25 & 24.74 \\
         & LFR & 70.02 & 95.55 & Adversarial & 64.49 & 3.56 & 3.95\\
         & Adversarial & 73.09 & 89.35 & Threshold Optimizer & 73.66 & 2.08 & 3.19\\
         & ROC & 69.83 & 95.77 & EGR & 62.94 & 2.26 & 2.43 \\ 
         & EGR & 62.06 & 95.98 &&&& \\ \hline \hline

         \multirow{4}{*}{\rotatebox[origin=rB]{90}{Credit}} &
         Biased & 81.58 & 85.28 & Biased & 81.58 & 1.16 & 1.06\\
        & LFR & 80.56 & 98.62 & Adversarial & 80.96 & 0.69 & 0.76\\ 
        & Adversarial &  80.48 & 99.86 & Threshold Optimizer & 81.25 & 0.16 & 1.06 \\
        & ROC & 81.58 & 98.99 & EGR & 75.14 & 1.39 & 0.2 \\ 
        & EGR & 72.84 & 90.69 &&&& \\ \hline \hline

        \multirow{4}{*}{\rotatebox[origin=rB]{90}{Dutch}} &
        Biased & 82.94 & 45.86 & Biased & 82.94  & 24.73 & 9.54\\
        & LFR & 73.59 &  84.65 & Adversarial & 74.02 & 0.73 & 0.27 \\
        & Adversarial & 74.15 & 92.34 & Threshold Optimizer & 79.62 & 0.62 & 0.11\\
        & ROC & 74.94 & 92.18 & EGR & 67.48 & 5.7  & 14.02 \\
        & EGR & 67.48 & 97.11 &&&& \\ \hline
         
    \end{tabular}}
    \caption{Performance Metrics of Biased and Fair Models across test sets of all Datasets.}
    \label{tab:metrics}
\end{table*}

\subsection{Final $E(\hat{Y})$ and FPR/TPR rates}

\begin{table*}[]
    \centering
    \begin{tabular}{|c|c|c|c|c|c|}
        \cline{2-6}
        \multicolumn{1}{c|}{}& Sensitive Attribute & LFR & Adversarial (DP) & ROC & Biased Model\\
        \hline
        \multirow{ 2}{*}{Adult} & Male & 17.43\% & 14.48\% & 9.32\% & 27.89\% \\
        \cline{2-6}
        & Female & 17.87\% & 14.12\% & 8.77\% & 8.77\% \\
        \hline
        \hline
        \multirow{ 2}{*}{Dutch} & Male & 42.87\% & 38.45\% & 30.74\% & 65.26\% \\
        \cline{2-6}
        & Female & 43.52\% & 36.70\% & 29.79\% & 29.79\%\\
        \hline
        \hline
        \multirow{ 2}{*}{COMPAS} & White & 54.54\% & 60.79\% & 41.85\% & 74.33\% \\
        \cline{2-6}
        & Black & 60.32\% & 56.72\% & 39.73\% & 39.73\%\\
        \hline
        \hline
        \multirow{ 2}{*}{Bank} & Male & 9.31\% & 7.77\% & 7.72\% & 15.25\% \\
        \cline{2-6}
        & Female & 7.86\% & 6.61\% & 6.71\% & 6.71\%\\
        \hline
        \hline
        \multirow{ 2}{*}{Credit} & Male & 11.06\% & 10.64\% & 11.74\% & 17.26\% \\
        \cline{2-6}
        & Female & 10.74\% & 10.44\% & 11.6\% & 14.8\%\\
        \hline
    \end{tabular}
    \caption{Percentages of subgroups classified into desirable class ($E(\hat{Y})$) by fair models achieved by each debiasing strategy optimizing Demographic Parity}
    \label{tab:eys}
\end{table*}

\begin{table*}[]
    \centering
    \begin{tabular}{|c|c|c|c|c|}
        \cline{2-5}
        \multicolumn{1}{c|}{}& Metric & Adversarial (EO) & Threshold Optimizer & Biased Model\\
        \cline{2-5}
        \hline
        \multirow{ 4}{*}{Adult} & \multirow{ 2}{*}{FPR} & Male: $9$\% & Male: 7\% & Male: 11\%  \\
        && Female: $6$\%  & Female: 5\% & Female: 2\%\\
        \cline{2-5}
        & \multirow{ 2}{*}{TPR} & Male: $55$\% & Male: 54\% & Male: 65\%  \\
        && Female: $63$\%  & Female: 57\% & Female: 57\% \\
        \hline
        \hline
        \multirow{ 4}{*}{Dutch} & \multirow{ 2}{*}{FPR} & 
        Male: $30$\%  & Male: 10\% & Male: 32\% \\
        && Female: $28$\% & Female: 10\% & Female: 7\% \\
        \cline{2-5}
        & \multirow{ 2}{*}{TPR} & 
        Male: $80$\% & Male: 68\%  & Male: 85\%  \\
        && Female: $81$\% & Female: 69\% & Female: 75\% \\
        \hline
        \hline
        \multirow{ 4}{*}{COMPAS} & \multirow{ 2}{*}{FPR} &
        White: $53$\% & White: 31\% & White: 46\%\\
        && Black: $50$\%  & Black: 30\% & Black: 21\% \\
        \cline{2-5}
        & \multirow{ 2}{*}{TPR} & 
        White: $80$\% & White: 78\% & White: 88\%  \\
        &&  Black: $78$\% & Black: 77\% & Black: 63\% \\
        \hline
        \hline
        \multirow{ 4}{*}{Bank} & \multirow{ 2}{*}{FPR} & 
        Male: $7$\%  & Male: 12\% & Male: 7\% \\
        &&  Female: $5$\% & Female: 8\% & Female: 3\%  \\
        \cline{2-5}
        & \multirow{ 2}{*}{TPR} & 
        Male: $50$\% & Male: 73\% & Male: 56\%\\
        && Female: $50$\%  & Female: 70\% & Female: 45\% \\
        \hline
        \hline
        \multirow{ 4}{*}{Credit} & \multirow{ 2}{*}{FPR}  &
        Male: $5$\% & Male: 7\% & Male: 8\% \\
        && Female: $5$\%  & Female: 7\% & Female: 7\%\\
        \cline{2-5}
        & \multirow{ 2}{*}{TPR}  &
        Male: $34$\% & Male: 39\% & Male: 43\% \\
        && Female: $32$\% & Female: 40\% & Female: 42\% \\
        \hline
    \end{tabular}
    \caption{TPR and FPR values ($E(\hat{Y}|y)$) by fair models achieved by each debiasing strategy optimizing Equalized Odds}
    \label{tab:eyys}
\end{table*}

\subsection{Impact Size Experiments for Gradient Boosting}

\begin{table*}[ht!]
    \centering
    \resizebox{0.9\textwidth}{!}
    {\begin{tabular}{|c|c|c|c||c|c|c|c|}
    \cline{2-8}
    \multicolumn{1}{c|}{} & \multicolumn{3}{c||}{Demographic Parity} & \multicolumn{4}{c|}{Equalized Odds} \\ 
    \cline{2-8} 
         \multicolumn{1}{c|}{} & Model & Accuracy [\%] & p\%-rule [\%] & Model & Accuracy [\%] & $D_{FPR}$ & $D_{TPR}$ \\ \cline{2-6} \hline
         \multirow{4}{*}{\rotatebox[origin=rB]{90}{Adult}} &
         Biased & $86.56$ & $29.30$ & Biased & $86.56$ & 5.90 & 9.34\\
         & LFR & 81.18 & $80.02$ & Threshold Optimizer & 79.53 & 2.41 & 3.06\\
         & ROC &  82.97  & $80.57$ & EGR & 86.10 & 1.93 & 4.62 \\  
         & EGR & 84.62 & 95.84 &&&& \\
         \hline \hline
         
         \multirow{4}{*}{\rotatebox[origin=rB]{90}{COMPAS}} &
         Biased & 74.19 & 62.15 &  Biased & 74.19 & 16.66 & 16.82 \\
         & LFR & 71.04 & 99.25 & Threshold Optimizer & 73.55 & 3.93 & 4.74\\
         & ROC & 70.10 & 97.04 & EGR & 71.61 & 10.18 & 10.14 \\ 
         & EGR & 66.91 & 80.08 &&&& \\ \hline \hline

        \multirow{4}{*}{\rotatebox[origin=rB]{90}{Dutch}} &
        Biased & 81.32 & 70.45 & Biased & 81.32 & 0.49 & 10.93 \\
        & LFR & 77.20 &  96.99 & Threshold Optimizer & 79.45 & 0.02 & 0.49 \\
        & ROC & 76.80 & 97.85 & EGR & 78.68 & 2.10 & 3.11 \\
        & EGR & 73.86 & 93.27 &&&& \\ \hline
         
    \end{tabular}}
    \caption{Performance Metrics of Biased and Fair Models across test sets of all Datasets, for Biased Model XGBoost.}
    \label{tab:metrics_xgb}
\end{table*}

\begin{table}[t]
   \centering
   \resizebox{\columnwidth}{!}
   {\begin{tabular}{|c|c|c|c|c|}
       \hline
       Dataset & Metric & Adversarial (EO) & TO (EO) & EGR (EO) \\ \hline \hline
       \multirow{2}{*}{Adult}& Mean $\%$ of instances changed & 7.42&4.61&16.42 \\
       & $\%$ instances changed in every run &1.27&1.56&0.06 \\ \hline \hline
       \multirow{2}{*}{Bank}& Mean $\%$ of instances changed & 9.57 &1.76& 9.62\\
       & $\%$ instances changed in every run & 0.88 &0.15& 0.13\\ \hline \hline
       \multirow{2}{*}{COMPAS}& Mean $\%$ of instances changed &25.12&11.45&29.92 \\
       & $\%$ instances changed in every run &5.26&9.58& 0.07\\ \hline \hline
       \multirow{2}{*}{Credit}& Mean $\%$ of instances changed &5.97&1.58& 17.37\\
       & $\%$ instances changed in every run &1.04&0.36&0.05 \\ \hline \hline
       \multirow{2}{*}{Dutch}& Mean $\%$ of instances changed &16.46&13.34&23.02 \\
       & $\%$ instances changed in every run &2.26&9.57& 0.01\\ \hline
   \end{tabular}}
   \caption{Stability of the instances targeted by EO methods across 10 runs. All results are in $\%$ of $\mathcal{X}_{val}$.}
    \label{tab:stability-eo}
\end{table}

\end{document}